\documentclass{article}

\usepackage[numbers]{natbib}
\usepackage[preprint]{neurips_2026}

\usepackage[utf8]{inputenc} 
\usepackage[T1]{fontenc}    
\usepackage{hyperref}       
\usepackage{url}            
\usepackage{booktabs}       
\usepackage{amsfonts}       
\usepackage{nicefrac}       
\usepackage{microtype}      
\usepackage{xcolor}         

\usepackage{wrapfig}
\usepackage[dvipsnames]{xcolor} 

\usepackage{colortbl} 
\usepackage{booktabs,multirow}
\usepackage{graphicx}
\usepackage{booktabs}
\usepackage[table]{xcolor}
\usepackage{bbm}   
\usepackage{caption}

\usepackage[accsupp]{axessibility}  

\title{Taming Outlier Tokens in Diffusion Transformers}

\author{%
  {
  Xiaoyu Wu\textsuperscript{1*}\quad
  Yifei Wang\textsuperscript{1*}\quad
  Tsu-Jui Fu\textsuperscript{2}\quad
  Liang-Chieh Chen\textsuperscript{2}\quad
  Zhe Gan\textsuperscript{2}\quad
  Chen Wei\textsuperscript{1}
  }
  \\[0.6em]
  \textsuperscript{1}Rice University
  \qquad
  \textsuperscript{2}Apple
}

\begin{document}

\maketitle
\renewcommand{\thefootnote}{*}
\footnotetext{Equal Contribution}
\renewcommand{\thefootnote}{\arabic{footnote}}
\begin{abstract}
We study outlier tokens in Diffusion Transformers (DiTs) for image generation. Prior work has shown that Vision Transformers (ViTs) can produce a small number of high-norm tokens that attract disproportionate attention while carrying limited local information, but their role in generative models remains underexplored. We show that this phenomenon appears in both the encoder and denoiser of modern Representation Autoencoder (RAE)-DiT pipelines: pretrained ViT encoders can produce outlier representations, and DiTs themselves can develop internal outlier tokens, especially in intermediate layers. Moreover, simply masking high-norm tokens does not improve performance, indicating that the problem is not only caused by a few extreme values, but is more closely related to corrupted local patch semantics. To address this issue, we introduce \emph{Dual-Stage Registers (DSR)}, a register-based intervention for both components:   trained registers when available, recursive test-time registers otherwise, and diffusion registers for the denoiser. Across ImageNet and large-scale text-to-image generation, these interventions consistently reduce outlier artifacts and improve generation quality. Our results highlight outlier-token control as an important ingredient in building stronger DiTs.

\vspace{-5pt}
\end{abstract}

\section{Introduction}

Vision Transformers (ViTs)~\cite{dosovitskiy2020image} have emerged as a foundational architecture in computer vision. Initially, they demonstrated strong performance on visual understanding tasks, including supervised classification (e.g., ViT~\cite{dosovitskiy2020image}, DeiT~\cite{touvron2021training}), self-supervised representation learning (e.g., DINO models~\cite{caron2021emerging,oquab2023dinov2,simeoni2025dinov3}), and contrastive language-image pretraining (e.g., CLIP~\cite{radford2021learning}, SigLIP~\cite{zhai2023sigmoid,tschannen2025siglip}). These successes gradually shifted the default backbone for visual recognition models from convolutional networks such as ResNets~\cite{he2016deep} to Vision Transformers~\cite{chen2022vision,raghu2021vision}.

More recently, ViTs have shown remarkable effectiveness in generative modeling, particularly for diffusion models~\cite{sohl2015deep,ho2020denoising,song2019generative}. Starting from DiT~\cite{peebles2023scalable}, Transformer-based generators have increasingly replaced UNet architectures~\cite{ho2020denoising,song2020score,rombach2022high} in both latent-space~\cite{ma2024sit,esser2024scaling} and pixel-space diffusion models~\cite{li2025back}. Beyond the generator itself, recent work such as RAE~\cite{zheng2025diffusion} further proposes using pretrained ViT encoders to project images into latents, replacing the UNet-based VAE~\cite{rombach2022high}, the last remaining convolutional component in state-of-the-art diffusion pipelines.

A commonly cited advantage of Transformers is their ability to model global interactions and their favorable optimization properties, albeit at the cost of weaker inductive biases. Recent studies, however, show that this flexibility can introduce nontrivial artifacts in the feature maps. In particular, Vision Transformers Need Registers~\cite{darcet2023vision} identifies the ubiquitous presence of outlier tokens in supervised and self-supervised ViTs (e.g., DeiT-III~\cite{touvron2022deit}, OpenCLIP~\cite{ilharco_gabriel_2021_5143773}, DINOv2~\cite{oquab2023dinov2}). These tokens exhibit abnormally large norms, absorb a disproportionate amount of attention, and yet carry little semantic information, resulting in artifacts in feature maps that hinder downstream utilization, especially for dense visual tasks. This behavior is closely related to the attention sink hypothesis in the context of large language models~\cite{xiaoefficient}, and can be effectively mitigated by introducing register tokens, i.e., dedicated non-patch tokens that explicitly serve as attention sinks during training.

However, investigations of outlier tokens have thus far focused almost exclusively on recognition models, leaving their role in generative models largely unexplored. This gap is particularly striking given the growing prevalence of Transformers in generative modeling and the heightened importance of locality and spatial fidelity in generation compared to recognition.
In this work, we show that outlier token effects are pervasive in both ViT-based autoencoder and diffusion Transformers.

For ViT-based autoencoders, we focus on RAE~\cite{zheng2025diffusion}, which has pretrained representation encoders. We empirically confirm the existence of outlier tokens in the encoders and demonstrate their negative impact on generation quality. Using the official DINOv2 implementation~\cite{oquab2023dinov2,darcet2023vision}, we compare DINOv2 encoders with and without register tokens and observe consistent improvements when registers are enabled. For SigLIP 2~\cite{tschannen2025siglip}, which does not include registers by default, we introduce recursive test-time registers~\cite{jiang2025vision} and show that this modification, too, leads to improved generation.

For Transformer-based generators, identifying outlier tokens is more subtle. In supervised and self-supervised ViTs, inputs and outputs differ in semantic abstraction, and prior work reports that outlier tokens typically emerge in late layers. In contrast, diffusion models operate on inputs and outputs of the same modality—local pixels or latents. Surprisingly, we find that outlier tokens still arise, but predominantly in intermediate layers of the diffusion Transformer. By introducing register tokens into the generator, we achieve consistent gains across a variety of diffusion architectures, including SiT~\cite{ma2024sit}, JiT~\cite{li2025back}, and RAE-based~\cite{zheng2025diffusion,tong2026scaling} designs. We refer to this unified use of register tokens in both the representation encoder and the diffusion generator as \emph{Dual-Stage Registers} (DSR).

Overall,  our results show that DSR provide a simple yet effective way to stabilize Transformer-based diffusion pipelines by addressing outlier tokens in both the tokenizer and the denoiser. This unified intervention consistently improves generation quality across settings. Concretely, for RAE-DiT with SigLIP2-B, it reduces ImageNet-256 FID from 5.89 to 4.58 and improves GenEval from 0.426 to 0.466 on a large-scale text-to-image task.

\section{Related Work}
\paragraph{Outlier Tokens in Transformers.}

In Vision Transformers, Darcet et al. identify a small fraction of patch tokens with unusually large feature norms, referred to as \emph{outlier tokens} \cite{darcet2023vision}. These tokens often carry limited local patch information while inducing irregular attention patterns. To mitigate this effect, they introduce extra learnable register tokens during training, which absorb the outlier behavior and lead to cleaner attention over image patches \cite{darcet2023vision}. Building on this finding, subsequent work traces these outliers to a sparse set of register neurons and proposes a training-free alternative: at inference time, the corresponding activations can be shifted into an additional untrained token, recovering much of the benefit of trained registers in models released without them \cite{jiang2025vision}.

Related sink-like phenomena have also been studied in language models. In this setting, \emph{attention sinks} attract disproportionate attention and have been linked to large residual-stream activations, motivating neuron-level or activation-level test-time interventions \cite{xiaoefficient,sun2024massive,yona2025interpreting}. More recent work  explores the connection between sink behavior and outlier-driven dynamics \cite{qiu2026unified}. Together, these studies suggest that outlier or sink tokens are a recurring phenomenon in Transformer architectures, and register-augmented ViT backbones have  become increasingly common in practice \cite{darcet2023vision}. By contrast, in Diffusion Transformers (DiTs), it remains unclear whether similar outlier tokens emerge during denoising and whether they affect generation quality. Our work addresses this gap.

\paragraph{Diffusion Transformers Across Different Input Spaces.}

DiTs are originally trained in variational autoencoder (VAE) latent spaces for efficiency, but the tokenizer can bottleneck reconstructed image quality because its latents may not fully preserve semantic information \cite{rombach2022high,peebles2023scalable}. Subsequent work improves latent DiT pipelines through representation alignment with visual features from vision foundation models, improving training efficiency and generation quality \cite{yu2024representation}. More recent work further enables end-to-end joint optimization of the VAE tokenizer and diffusion model \cite{leng2025repa}.
 
 Apart from VAE space, RAE-based pipelines replace VAEs with pretrained representation encoders paired with learned decoders, yielding semantically richer latents and faster convergence in DiT training \cite{zheng2025diffusion}. In parallel, another line of work revisits pixel-space diffusion to remove the need of tokenizers, motivated by the desire to avoid lossy encode--decode stages and simplify the modeling stack. These approaches train diffusion transformers directly on images and show competitive performance when paired with appropriate architectures and training recipes \cite{li2025back,baade2026latent}.




\section{Outlier Tokens in RAE-DiT}

\begin{figure}[t]
    \centering
    \includegraphics[width=\linewidth]{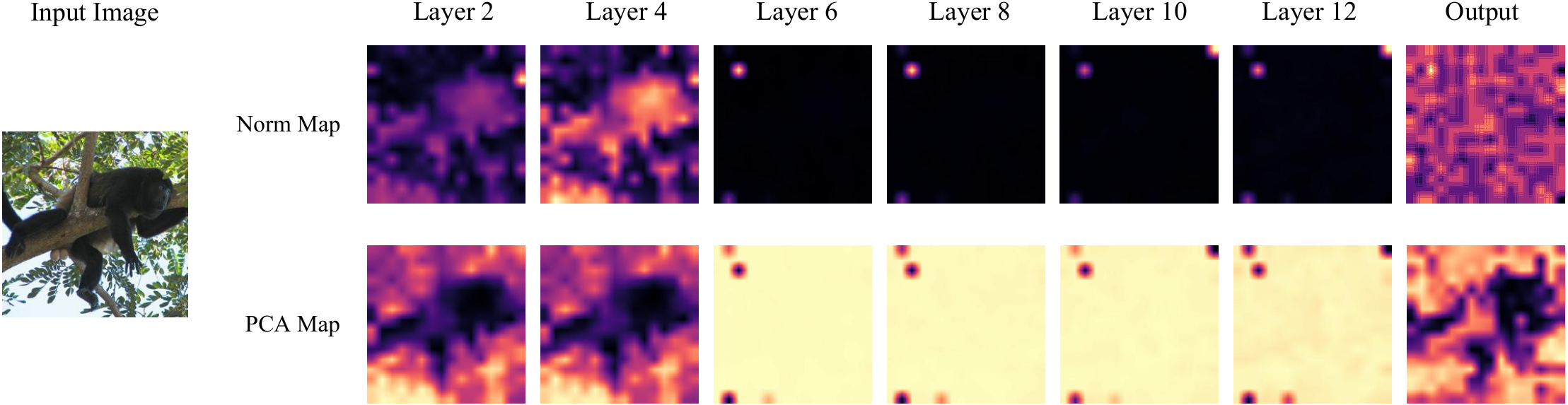}
\caption{\textbf{Outlier tokens in ViT-based autoencoders.}
We visualize token-norm maps across the layers of the SigLIP2-B encoder. Severe high-norm tokens emerge in the last few layers: the penultimate layer shows the strongest outlier pattern, while the final output becomes somewhat more stable, potentially due to the reconstruction-related training objective in SigLIP2.}
    \label{fig:encoder_outlier}
    \vspace{-14pt}
\end{figure}

RAE-DiT consists of two Transformer-based stages: a ViT-based representation encoder that serves as the tokenizer, and a diffusion Transformer that generates in the representation space.  We examine outlier tokens in both stages and find distinct outlier patterns in the encoder and the diffusion generator.

\subsection{Outlier Tokens in ViT-based Autoencoders}
\label{sec:encoder_outliers}

Outlier tokens have been observed in standard ViTs, where they typically appear in the final layers and can be mitigated by register tokens~\cite{darcet2023vision}. In RAE-DiT, such outliers in encoders are especially relevant because the encoder features define the representation space used for diffusion training.

As shown in Fig.~\ref{fig:encoder_outlier}, using SigLIP2-B as an example, we observe severe high-norm tokens in the last few encoder layers. The penultimate layer exhibits the strongest outliers, while the final output becomes more stable, potentially due to the reconstruction-related training objective in SigLIP2. These observations suggest that outlier tokens already exist in the encoders before diffusion training.

\subsection{Outlier Tokens in Transformer-based Generators}

\begin{figure}[t]
    \centering
    \includegraphics[width=\linewidth]{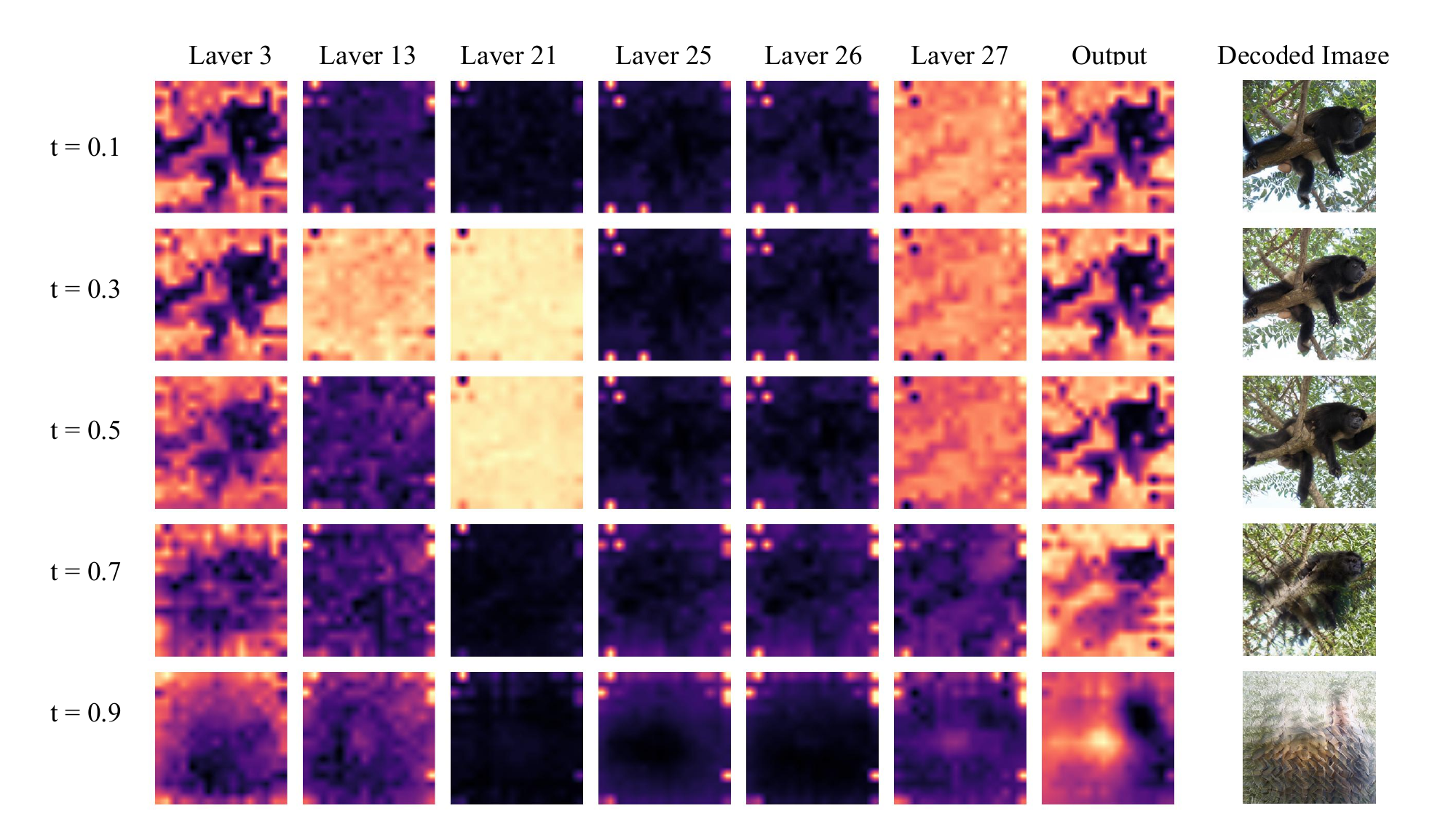}
\caption{\textbf{Outlier Tokens in Transformer-based Generators.} We visualize token-norm maps of RAE-DiT with a SigLIP2-B encoder, across different diffusion noise scales and encoder layers. We find that high-norm outliers concentrate in the \emph{intermediate} layers, while their severity decreases as the diffusion noise level increases. This pattern differs from prior observations in standard ViTs, where artifact tokens are typically most pronounced in the final layers.}
    \label{overview_outlier}
\end{figure}

Beyond prior analyses of vision foundation models (standard ViTs), we identify a consistent outlier-token phenomenon in the RAE-DiT pipeline: a small subset of tokens in the representation encoder exhibits abnormally large norms, and this effect is further amplified when the encoder features are used as conditioning for diffusion training. As shown in Fig.~\ref{overview_outlier}, for RAE-DiT with SigLIP2-B as the encoder, outliers concentrate in \emph{intermediate} layers, and their severity decreases as the diffusion noise level increases, suggesting a compounding mechanism in which encoder anomalies are amplified by the denoising objective. This differs from the standard ViT pattern discussed above, where outliers are typically most pronounced in the \emph{final} layers~\cite{darcet2023vision}.

To test whether the degradation is merely an \emph{extreme-value} effect—namely, whether a small number of large-loss tokens dominate optimization—we apply a simple token-level loss masking strategy in representation-space diffusion training. Let $z_0 \in \mathbb{R}^{N \times d}$ denote the clean token representation, and let $z_t = \alpha(t) z_0 + \sigma(t)\epsilon$ be the noisy tokens, where $\epsilon \sim \mathcal{N}(0, I)$ and $t \in [0,1]$. Following prior work~\cite{li2025back}, the generator predicts $\hat z_0 = x_\theta(z_t, t, c)$, which we convert to a $v$-prediction as $\hat v_\theta = (\hat z_0 - z_t)/(1-t)$; the corresponding target is $v = (z_0 - z_t)/(1-t)$.

We then mask tokens according to their representation norms, defining
\[
m_i = \mathbf{1}\!\left[\|z_{0,i}\|_2 \le \tau\right].
\]
The masked training objective is
\begin{equation}
\mathcal{L}_{v}^{\mathrm{mask}}(\theta)
= \mathbb{E}_{z_0,\epsilon,t}\!\left[
\frac{1}{\sum_i m_i}\sum_{i=1}^{N}
m_i \left\| \hat{v}_\theta(z_t,t,c)_i - v_i\right\|_2^2
\right]
.
\label{eq:v_loss_masked}
\end{equation}
In other words, tokens with unusually large representation norms are excluded from the diffusion loss. If the degradation were driven primarily by a small number of extreme-loss tokens, this masking strategy would be expected to substantially mitigate the problem.

\begin{table}[t]
\vspace{-14pt}
\centering
\small
\setlength{\tabcolsep}{5.5pt}
\renewcommand{\arraystretch}{1.15}
\caption{Effects of loss masking.}
\begin{tabular}{lccccc}
\toprule
Training Strategies &\% of tokens filtered & FID$\downarrow$ & IS$\uparrow$ & Prec.$\uparrow$ & Rec.$\uparrow$ \\
\midrule

RAE-DiT-XL (SigLIP2-B)                         & 0\% & 5.89 & 156.54		 & 0.686 & 0.562 \\
+ loss masking ($\tau=100$)                   &  0.1\%    &  6.06 & 152.72 & 0.686 & 0.562 \\


\bottomrule
\end{tabular}
\vspace{-14pt}
\label{tab:training_strategies_sigliP2-B-loss-mask}
\end{table}

This method simply discards the training signal for tokens identified as encoder outliers. However, as shown in Tab.~\ref{tab:training_strategies_sigliP2-B-loss-mask}, masking does not improve generation quality, indicating that the degradation cannot be explained primarily by a few extreme loss values. Instead, we hypothesize that outlier tokens are a \emph{symptom} of corrupted local patch information: removing supervision at those positions cannot restore the missing patch-level semantics, and may instead further weaken local learning.

This view is supported by prior findings that generation quality depends on preserving local spatial structure in the representation space. Analyses in iREPA suggest that patchwise structure, rather than global semantics alone, is closely tied to generative performance \cite{singh2025matters}. Studies of ViT feature artifacts show that high-norm or outlier tokens can harm dense prediction tasks such as segmentation, which depend on accurate patch semantics \cite{darcet2023vision}. Together, these observations motivate us to focus in the following section on restoring patch-level semantics, rather than suppressing extreme values.

\section{Improving Generation with Dual-Stage Registers (DSR)}

\begin{figure}[t]
    \centering
    \includegraphics[width=\linewidth]{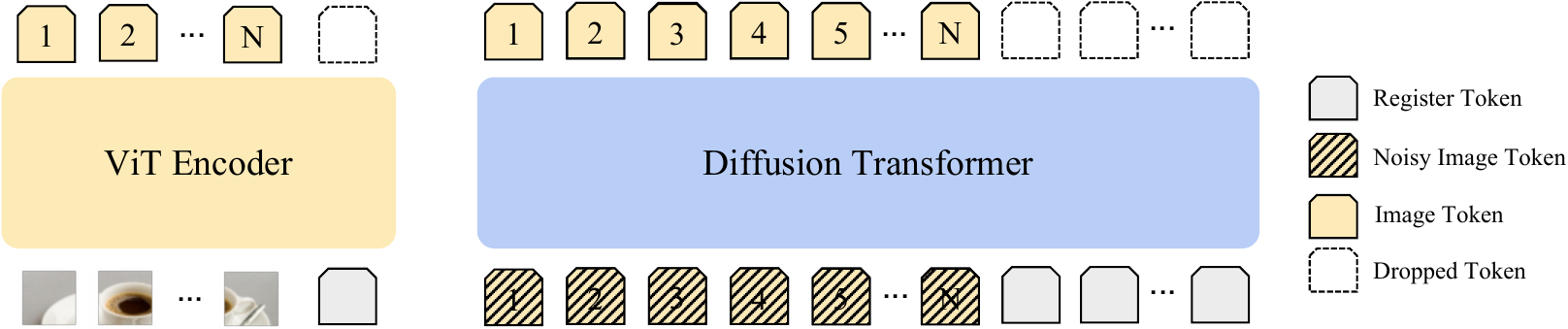}
\caption{\textbf{Framework of our Dual-Stage Registers (DSR) method.}
Our DSR method patches both the vision encoder and the diffusion model with register tokens. The encoder uses a \emph{test-time register token}, which is inserted only at inference time, while the diffusion model uses 36 \emph{trained register tokens}, which are learned during diffusion training. During training, we discard the encoder-side register-token outputs before feeding encoder features into the diffusion model. During inference, we discard register-token outputs from both modules and keep only the image-token outputs.}

    \label{fig:framework}
\end{figure}

The results above suggest that the degradation cannot be explained solely by a few extreme token losses. Instead, a more plausible explanation is that outlier tokens reflect degraded local patch semantics and spatial structure in the representation space. This interpretation is consistent with prior findings that local structure is important for generative quality \cite{singh2025matters}, and with analyses of artifact tokens in ViTs that connect such anomalies to failures in dense prediction \cite{darcet2023vision}. Motivated by this view, we introduce \emph{Dual-Stage Registers} (DSR), a lightweight intervention that absorbs token-level artifacts and stabilizes patch representations across the RAE-DiT pipeline. As shown in Fig.~\ref{fig:framework}, DSR patches \emph{both} sides of the pipeline: the vision encoder and the diffusion transformer.

\subsection{Registers in Vision Encoders}

\begin{figure}[t]
    \centering
    \includegraphics[width=\linewidth]{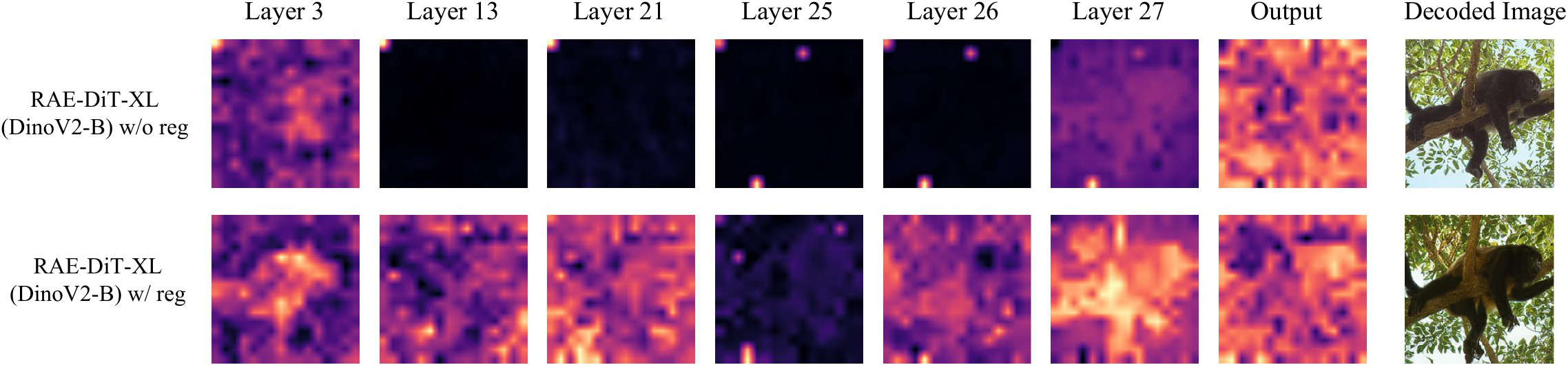}
\caption{\textbf{Norm map comparison for validating trained encoder registers.} We compare RAE-DiT(DINOv2-B) w/o and w/  trained encoder registers, at a fixed timestep $t=0.5$. We find that introducing trained registers consistently suppresses high-norm token outliers and improves the quality of patch-level representations, which in turn leads to stronger downstream generation.}

\vspace{-0.2in}
    \label{fig:enc_reg_main}
\end{figure}

We begin by validating the effect of \emph{trained} registers in a vision encoder where such registers are available. Specifically, we compare DINOv2~\cite{oquab2023dinov2} trained with and without register tokens\footnote{
For this comparison, we use the released DINOv2 checkpoints with and without register tokens.}, measuring outlier severity and downstream generation quality when a DiT is trained on the resulting representation space. As shown in Fig.~\ref{fig:enc_reg_main} and Tab.~\ref{tab:training_strategies_dino}, trained registers  reduce high-norm token artifacts and improve the quality of patch-level representations, leading to stronger generative performance.

\begin{table}[h]
\centering
\small
\setlength{\tabcolsep}{7pt}
\renewcommand{\arraystretch}{1.2}
\vspace{-0.2in}
\caption{Comparison of RAE-DiT (DINOv2-B) on ImageNet, w/ and w/o encoder register tokens.}

\begin{tabular}{lcccc}
\toprule
Training Strategies & FID$\downarrow$ & IS$\uparrow$ & Prec.$\uparrow$ & Rec.$\uparrow$ \\
\midrule

RAE-DiT-XL (DINOV2-B-no-reg)                          &4.16  & 203.46 & 0.787 & 0.525 \\
RAE-DiT-XL (DINOV2-B-with-reg)                          &  3.95 & 216.84 & 0.737 & 0.547 \\
\bottomrule
\end{tabular}
\label{tab:training_strategies_dino}
\end{table}

We next study SigLIP2~\cite{tschannen2025siglip}, a widely used vision-language encoder in recent multimodal models, to test whether our approach remains effective in a more practically important setting. In our experiments, the outlier phenomenon is substantially more severe for SigLIP2, while the off-the-shelf checkpoints do not provide trained registers. We therefore adopt \emph{test-time register} tokens (TTR), following prior work \cite{jiang2025vision}, by appending an additional token to the input sequence at inference time without further encoder training. Empirically, we find that the SigLIP2-So400 pipeline exhibits \emph{two} distinct sources of outliers, as discussed in Appendix Sec.~\ref{two-outlier}. To address this, we apply TTR \emph{recursively}: we first use TTR to stabilize the encoder output, and then apply it again to the resulting representation when a second-stage outlier pattern is detected. As shown in Tab.~\ref{tab:training_strategies_sigliP2-B}, this recursive TTR scheme consistently reduces outlier severity and improves generation quality.

\begin{figure}[t]
    \centering
    \includegraphics[width=\linewidth]{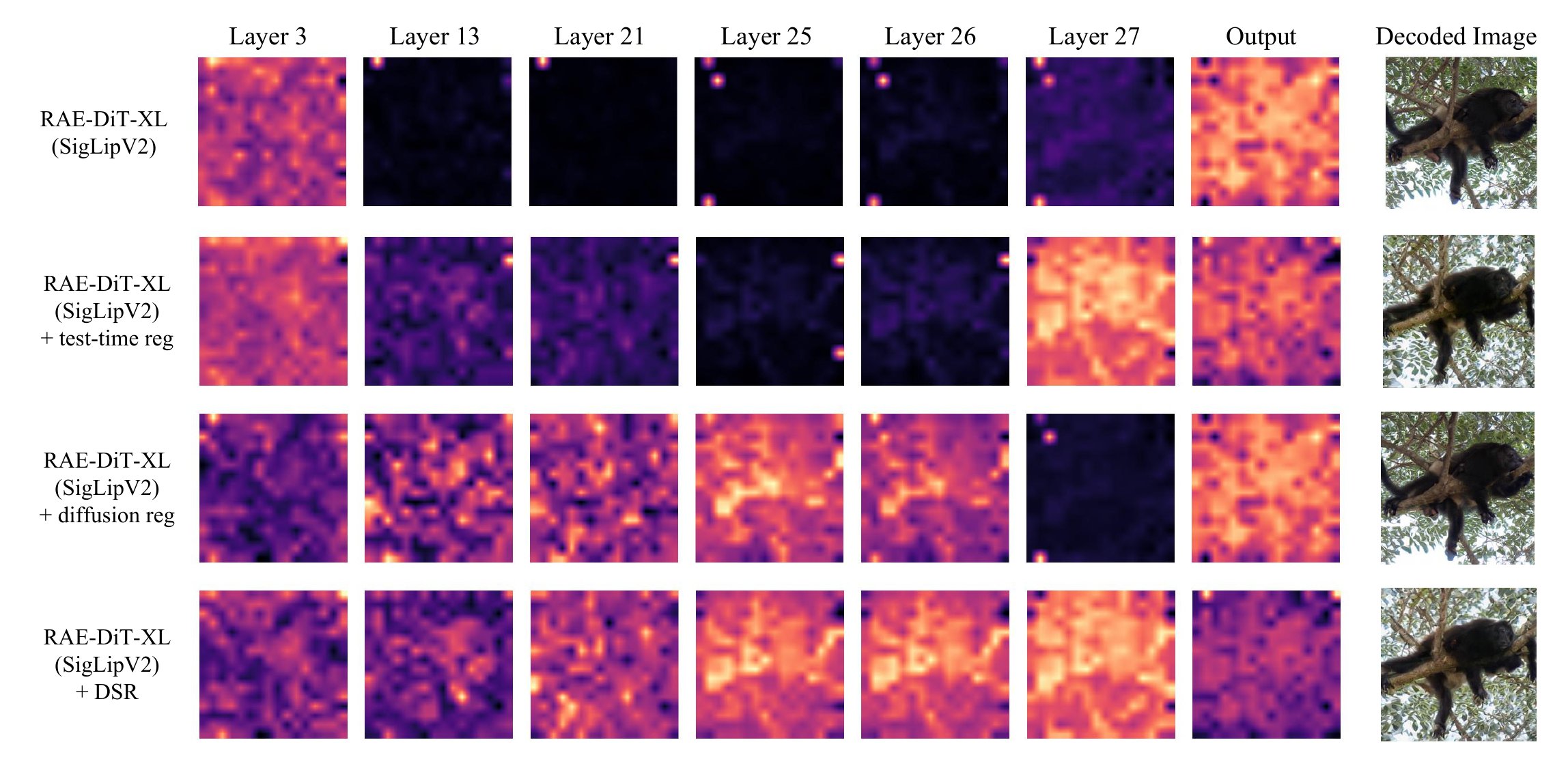}
    \caption{\textbf{Norm map comparison across variants.} We compare the baseline with two register-token configurations: adding test-time registers in the encoder only, and further adding trained registers in the diffusion model. We find that outliers in the norm map are suppressed only when \emph{both} sources of outliers are addressed, i.e., when registers are applied to both the encoder and the diffusion model.}
    \label{fig:norm_map}
    \vspace{-0.2in}
\end{figure}

\begin{figure}[t]
    \centering
    \includegraphics[width=\linewidth]{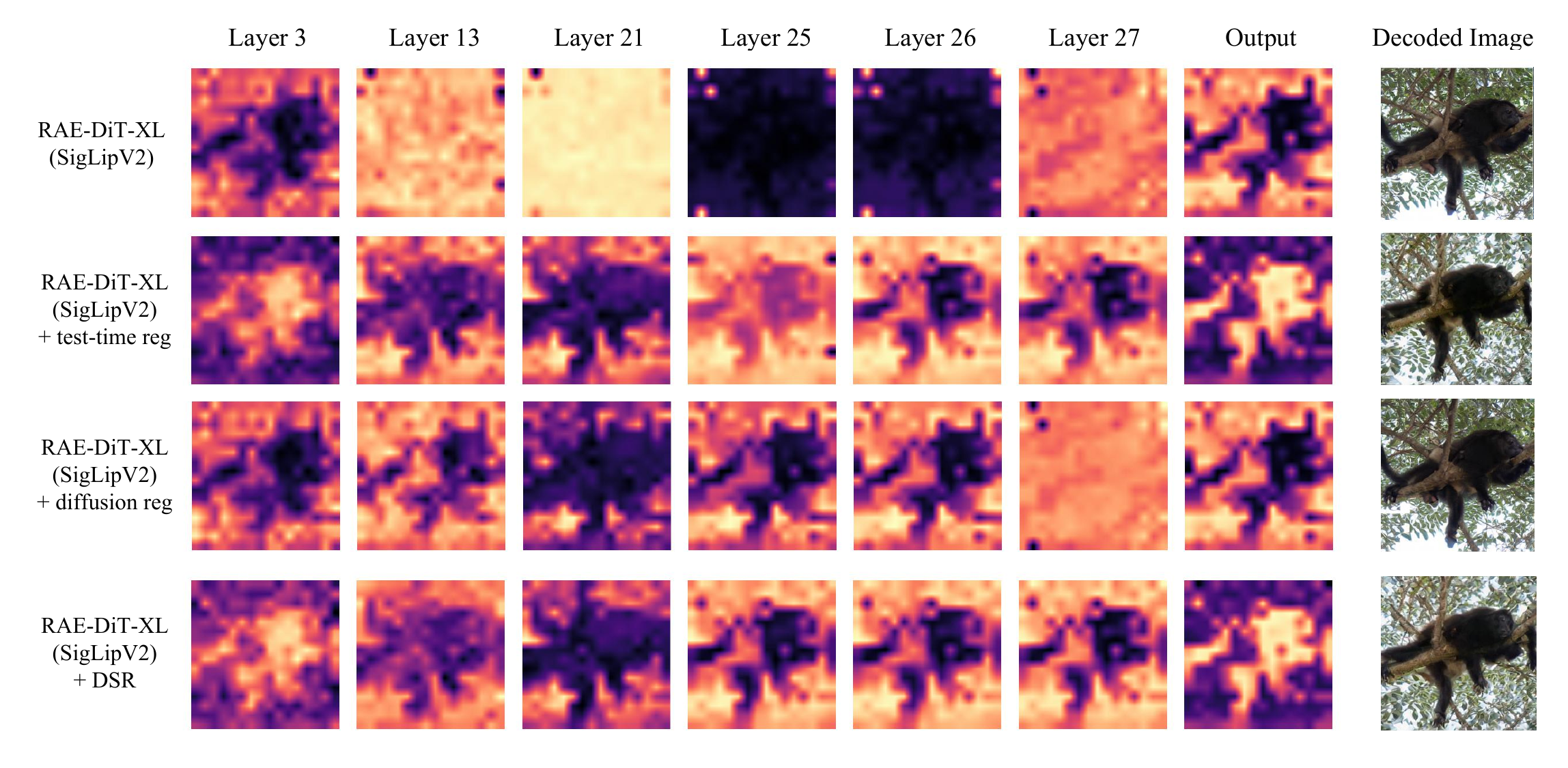}
    \caption{\textbf{PCA map comparison across variants.} We observe that adding test-time registers in the encoder yields a strong and visible improvement in the PCA map. Further adding trained registers in the diffusion model brings some additional improvements.}
    \label{fig:pca_map}
    \vspace{-0.2in}
\end{figure}

\begin{table}[t]
\centering
\small
\setlength{\tabcolsep}{7pt}
\renewcommand{\arraystretch}{1.2}
\caption{\textbf{Effect of diffusion registers across different DiT  on ImageNet.} Baseline rows are evaluated without diffusion registers; indented rows add diffusion registers on top of the same setting.}
\begin{tabular}{lcccc}
\toprule
Method & FID$\downarrow$ & IS$\uparrow$ & Prec.$\uparrow$ & Rec.$\uparrow$ \\

\midrule
RAE-DiT-XL (DINOv2-B, w/ encoder reg)           & 4.11 & 226.44 & 0.775 & 0.529 \\
\quad + diffusion reg                      & \textbf{3.92} & 226.92 & 0.773 & 0.542 \\
RAE-DiT-XL (SigLIP2-B)       & 5.89 & 156.54		 & 0.686 & 0.562\\
\quad + diffusion reg                      & \textbf{5.33} & 166.2 & 0.702 & 0.556 \\
RAE-DiT-XL (SigLIP2-B, w/ test-time reg)        & 4.63 & 177.2 & 0.748 & 0.542\\
\quad + diffusion reg                      & \textbf{4.58} & 165.99 & 0.725 & 0.56 \\
\midrule
VAE-SiT-XL                                    & 16.05 & 70.11 & 0.550 & 0.647 \\
\quad + diffusion reg                      & \textbf{14.47} & 78.50 & 0.554 & 0.651 \\
\midrule
JIT-H                                  & 30.34 & 22.34 & 0.424 & 0.621 \\
\quad + diffusion reg                       & \textbf{23.14} & 26.36 & 0.475 & 0.611 \\

\bottomrule
\end{tabular}

\label{tab:diffusion_registers_consistency}
\vspace{-14pt}
\end{table}

\begin{table}[t]
\centering
\small
\setlength{\tabcolsep}{7pt}
\renewcommand{\arraystretch}{1.2}
\caption{\textbf{Register tokens vs. in-context conditioning across DiTs and input spaces.} In-context conditioning yields smaller gains as the input representation becomes more semantic, whereas register tokens provide consistent improvements.}

\begin{tabular}{lcccc}
\toprule
Method & FID$\downarrow$ & IS$\uparrow$ & Prec.$\uparrow$ & Rec.$\uparrow$ \\

\midrule
RAE-DiT-XL (SigLIP2-B)       & 5.89 & 156.54		 & 0.686 & 0.562  \\
\quad + diffusion reg                      & \textbf{5.33} & 166.2 & 0.702 & 0.556 \\
\quad + diffusion in-context condition                      & 5.79 & 164.99 & 0.696 & 0.546 \\
\midrule
VAE-SiT-XL                                    & 16.05 & 70.11 & 0.550 & 0.647 \\
\quad + diffusion reg                      & 14.47 & 78.50 & 0.554 & 0.651 \\
\quad + diffusion in-context condition                      & \textbf{14.31} & 77.04 & 0.559 & 0.646 \\
\midrule
JIT-H                                  & 30.34 & 22.34 & 0.424 & 0.621 \\
\quad + diffusion reg                       & 23.14 & 26.36 & 0.475 & 0.611 \\
\quad + diffusion in-context condition                     & \textbf{15.51} & 33.05 & 0.519 & 0.618 \\

\bottomrule
\end{tabular}

\label{tab:incontext}
\vspace{-14pt}
\end{table}

\subsection{Registers in Diffusion Transformers}

While encoder-side registers substantially reduce representation outliers, we still observe persistent outlier tokens inside the diffusion transformer itself, especially in intermediate layers, as shown in Fig.~\ref{fig:norm_map}. This suggests that the generator can also develop a small set of high-norm tokens, potentially encouraged by the global aggregation behavior of self-attention and the need to maintain global context throughout the denoising process \cite{xiaoefficient}. Motivated by this observation, we introduce a small number of \emph{trainable diffusion registers} into the diffusion transformer. These registers are learned jointly with the generator and removed at inference time, following common practice in prior work on registers \cite{darcet2023vision,jiang2025vision}. With diffusion registers, the internal outlier pattern is largely eliminated, as shown in Fig.~\ref{fig:norm_map} and Fig.~\ref{fig:qual_map}. Intermediate-layer semantics are also improved, as illustrated by the PCA visualization in Fig.~\ref{fig:pca_map}, and generation quality improves consistently, as reported in Tab.~\ref{tab:diffusion_registers_consistency}.






More precisely, we find that diffusion registers yield consistent improvements across a wide range of settings. In particular, they improve performance for every input-space variant we evaluate, including pixel space, VAE latents, and multiple representation encoders. This suggests that the gains are not tied to any particular encoder or representation family. Instead, the broad effectiveness of diffusion registers indicates that they address token-level artifacts arising \emph{within} the diffusion transformer itself, rather than serving as a modification tailored to a specific representation.

We also observe a close relationship between diffusion registers and in-context conditioning~\cite{li2025back}. Both augment the input with extra tokens that can carry global context: in-context conditioning uses input-dependent tokens, while registers provide dedicated learnable slots. In Tab.~\ref{tab:incontext}, we find that in-context conditioning is most competitive when the input space contains less semantic or structured token representations, while diffusion registers remain beneficial as representations become more semantic. These results suggest that registers provide a simpler and more robust mechanism for supplying global capacity, without relying on the semantics or formatting of in-context tokens.







\section{Experiments}


\begin{figure}[t]
    \centering
    \begin{minipage}[t]{0.43\linewidth}
    \vspace{0pt}
    \centering
    \includegraphics[width=\linewidth]{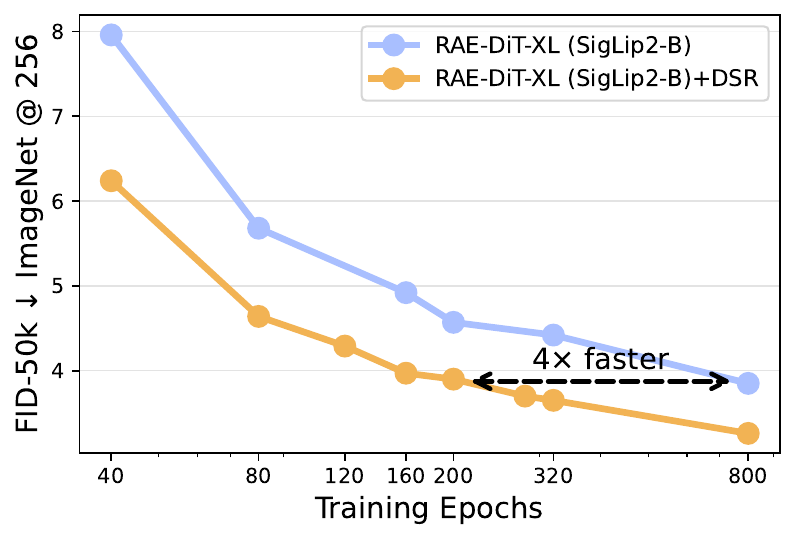}
        \vspace{-20pt}
        \caption{FID vs. epochs on IN-1K $256^2$.}
        \label{fig:fid_compare}
    \end{minipage}
    \hfill
    \begin{minipage}[t]{0.55\linewidth}
        \vspace{0pt}
        \centering
        \small
        \setlength{\tabcolsep}{2pt}
        \renewcommand{\arraystretch}{1.35}
        \captionof{table}{Test-time registers on RAE-DiT with SigLIP2-B and SigLIP2-So400, trained for 80 epochs.}\label{tab:training_strategies_sigliP2-B}
        \begin{tabular}{lcccc}
        \toprule
        Training Strategies & FID$\downarrow$ & IS$\uparrow$ & Prec.$\uparrow$ & Rec.$\uparrow$ \\
        \midrule
        RAE-DiT-XL (SigLIP2-B) & 5.89 & 156.54 & 0.686 & 0.562 \\
        + test-time register & 4.63 & 177.2 & 0.748 & 0.542 \\
        RAE-DiT-XL (SigLIP2-So400) & 7.04 & 167.01 & 0.682 & 0.515 \\
        + test-time register & 6.66 & 166.88 & 0.687 & 0.527 \\
        + test-time register (recursive) & 6.48 & 163.35 & 0.684 & 0.531 \\
        \bottomrule
        \end{tabular}
    \end{minipage}
\end{figure}

\begin{table*}[t]
\centering
\small
\setlength{\tabcolsep}{3pt}
\renewcommand{\arraystretch}{1.15}
\caption{Class-conditional performance on ImageNet 256$\times$256.}

\resizebox{\textwidth}{!}{%
\begin{tabular}{l c c c c c c c c c c}
\toprule
Method & Epochs & \#Params &
\multicolumn{4}{c}{Generation@256 w/o guidance} &
\multicolumn{4}{c}{Generation@256 w/ guidance} \\
\cmidrule(lr){4-7}\cmidrule(lr){8-11}
& & & gFID$\downarrow$ & IS$\uparrow$ & Prec.$\uparrow$ & Rec.$\uparrow$
    & gFID$\downarrow$ & IS$\uparrow$ & Prec.$\uparrow$ & Rec.$\uparrow$ \\
\midrule

\multicolumn{11}{l}{\textit{Pixel Diffusion}} \\
\arrayrulecolor{gray!50}\hline
\arrayrulecolor{black}
ADM~\cite{dhariwal2021diffusion}          & 400 & 554M & 10.94 & 101.0 & 0.69 & 0.63 & 3.94 & 215.8 & 0.83 & 0.53 \\
JiT-H/16 ~\cite{li2025back}    & 600  & 953M & --    & --    & --   & --   & 1.86 & 303.4 & --            & --   \\
\midrule

\multicolumn{11}{l}{\textit{Latent Diffusion}} \\
\arrayrulecolor{gray!50}\hline
\arrayrulecolor{black}

SiT-XL~\cite{ma2024sit}          & 1400 & 675M & 8.61 & 131.7 & 0.68 & 0.67 & 2.06 & 270.3 & 0.82          & 0.59 \\



\multirow{1}{*}{REPA~\cite{yu2024representation}}
             & 800 & 675M & 5.78 & 158.3 & 0.70 & 0.68 & 1.29 & 306.3 & 0.79          & 0.64 \\

\multirow{1}{*}{REPA-E~\cite{leng2025repa}}
             & 800 & 675M & 1.70 & 217.3 & 0.77 & 0.66 & 1.15 & 304.0 & 0.79          & 0.66 \\

\multirow{1}{*}{RAE-DiT$^{\scriptsize\mathrm{DH}}$-XL (DINOv2-B)}
             & 800 & 839M & 1.51 & 242.9 & 0.79 & 0.63 & 1.13 & 262.6 & 0.78 & 0.67 \\




\midrule
\multicolumn{11}{l}{\textit{Latent Diffusion with Multi-modal encoder}} \\
\arrayrulecolor{gray!50}\hline
\arrayrulecolor{black}
\multirow{2}{*}{RAE-DiT-XL (SigLiP2-B)} & 80& \multirow[c]{2}{*}{676M}  & 5.89 & 156.54 & 0.686 & 0.562   & -- & -- & -- & -- \\

 & 800 &  & 3.85 & 179.82 & 0.692 & 0.613 & 3.58 & 194.67 & 0.691 & 0.619 \\
\arrayrulecolor{gray!50}\hline
\arrayrulecolor{black}

\multirow{2}{*}{RAE-DiT-XL (SigLiP2-B) + DSR} & 80 & \multirow[c]{2}{*}{676M}& 4.58 & 165.99 & 0.725 & 0.56 & -- & -- & -- & -- \\

 & 800 &  & 3.26 & 185.54 & 0.704 & 0.62 & 2.97 & 203.95 & 0.709 & 0.621 \\

\arrayrulecolor{gray!50}\hline
\arrayrulecolor{black}

\multirow{2}{*}{RAE-DiT$^{\scriptsize\mathrm{DH}}$ (SigLiP2-B)} & 80& \multirow[c]{2}{*}{839M}  & 3.74 & 179.24 & 0.707 & 0.599  & -- & -- & -- & --  \\

 & 800 &  & 2.91 & 204.73 & 0.686 & 0.642 & 2.77 & 221.76 & 0.685 & 0.649 \\
\arrayrulecolor{gray!50}\hline
\arrayrulecolor{black}
\multirow{2}{*}{RAE-DiT$^{\scriptsize\mathrm{DH}}$-XL (SigLiP2-B) + DSR} & 80& \multirow[c]{2}{*}{839M}& 3.56 & 181.08 & 0.722 & 0.589 & -- & -- & -- & --  \\
 & 800 &  & 2.72 & 207.76 & 0.697 & 0.646 & 2.62 & 223.85 & 0.697 & 0.651 \\

\bottomrule
\end{tabular}}
\vspace{-0.1in}
\label{tab:scale}
\end{table*}

We evaluate DSR on ImageNet-1K class-conditional generation at $256\times256$ and text-to-image generation. For ImageNet-1K, we follow the RAE setting~\cite{zheng2025diffusion} and keep the same training epochs, learning rate, and model architecture for fair comparison; our main configuration uses SigLIP2-B as the encoder. While our encoder is slightly different due to the test-time register modification, we use the same decoder as in RAE; we find that retraining the decoder has limited impact (Appendix Sec.~\ref{retrained_decoder}). Additionally, we use data prediction with velocity loss to train our model and RAE baseline, which has been proven effective in JiT~\cite{li2025back}.

For ImageNet-1K, Tab.~\ref{tab:scale} shows that DSR substantially reduces gFID for \textsc{RAE} (SigLIP2-B). Combining DSR with the DDT head yields further gains and achieves competitive performance. Fig.~\ref{fig:fid_compare} plots the per-epoch trajectory and shows that DSR reaches comparable quality with $4\times$ fewer epochs.

\subsection{Scalability}

A key motivation for decoupling the Transformer design from the task is to better leverage scaling~\cite{li2025back, peebles2023scalable}. We reports ImageNet $256\times256$ results across model sizes (DiT-B/L/XL). As shown in Tab.~\ref{tab:scalability_sigliP2-B_100k}, DSR consistently improves gFID across all parameter scales, while introducing only a minor increase in GFLOPs.

\begin{table}[t]
\centering
\small
\setlength{\tabcolsep}{7pt}
\renewcommand{\arraystretch}{1.2}
\caption{Scalability of DSR across DiT model sizes for RAE-DiT with SigLIP2-B on ImageNet 256$\times$256. All models are trained for 100k iterations under the same training and evaluation protocol. For each size, we compare the baseline with DSR.}
\begin{tabular}{lccccc}
\toprule
Model  & FID$\downarrow$ & IS$\uparrow$ & Prec.$\uparrow$ & Rec.$\uparrow$& Gflops.$\downarrow$ \\
\midrule

RAE-DiT-S (SigLIP2-B)                 & 28.03 & 66.15 & 0.277 & 0.302  & 12.4 \\
\quad + DSR                    & 23.93 & 63.86 & 0.498 & 0.470 & 13.7$_{\textcolor{BrickRed}{+10.4\%}}$\\
\addlinespace[2pt]

RAE-DiT-B (SigLIP2-B)                 & 20.36 & 78.76 & 0.539 & 0.493 & 46.6  \\
\quad + DSR                    & 9.81 & 110.23 & 0.637 & 0.543 & 51.16$_{\textcolor{BrickRed}{+9.9\%}}$ \\
\addlinespace[2pt]


RAE-DiT-XL (SigLIP2-B)                & 5.89 & 156.54		 & 0.686 & 0.562 & 238.1\\
\quad + DSR                    & 4.58 & 165.99 &	0.725 &	0.560 & 262.9$_{\textcolor{BrickRed}{+10.4\%}}$\\

\bottomrule
\end{tabular}

\label{tab:scalability_sigliP2-B_100k}
\vspace{-5pt}
\end{table}

\begin{wraptable}{r}{0.46\textwidth}
\vspace{-12pt}
\centering
\caption{Text-to-image perf. on Scale-RAE.}
\begin{tabular}{lcc}
\toprule
\multicolumn{1}{c}{Models} &
GenEval$\uparrow$ &
DPG-Bench$\uparrow$ \\
\midrule
Baseline & 42.6 & 74.3 \\
Ours & 46.6 & 75.4 \\
\bottomrule
\end{tabular}
\vspace{-10pt}
\label{tab:scale-rae}
\end{wraptable}
We also demonstrate the scalability of DSR on Scale-RAE~\cite{tong2026scaling}, which leverages the MetaQuery architecture~\cite{pan2025transfer} for text-to-image (T2I) generation and unified modeling. We use SigLIP2-B~\cite{tschannen2025siglip} as the encoder and train the model on the Scale-RAE dataset~\cite{tong2026scaling}, which contains 24.7M synthetic images generated by FLUX.1-schnell~\cite{blackforestlabs_flux_2024}.

The training process takes 12 hours on a google cloud v5p-128 TPU for 10k training steps with a batch size of 2048. We then sample images from the trained model using text prompts and decode the generated latents with SigLIP-B pretrained decoder provided by RAE~\cite{zheng2025diffusion}. We evaluate the sampled images on GenEval~\cite{ghosh2023geneval} and DPG-bench~\cite{hu2024ella}.
As shown in Tab.~\ref{tab:scale-rae}, DSR achieves better performance than the original RAE baseline on both benchmarks. More details can be found in Appendix Sec.~\ref{sec:scale-rae}.

\subsection{Ablation Studies}

\begin{table}[t]
\centering
\small
\setlength{\tabcolsep}{6.5pt}
\renewcommand{\arraystretch}{1.15}
\caption{Ablation on diffusion registers for RAE-DiT-XL (SigLIP2-B) on ImageNet 256$\times$256. We vary the insertion starting block (with a fixed number of 36 registers) and the number of registers (with a fixed starting block of 8).}
\begin{tabular}{c c c c c c c}
\toprule
Starting block & \#Regs & FID$\downarrow$ & IS$\uparrow$ & Pre.$\uparrow$ & Rec.$\uparrow$ & Gflops $\downarrow$ \\
\midrule

 - & 0  & 5.89 & 156.54		 & 0.686 & 0.562 & 238.1 \\
\midrule

 0  & 36 & 5.54 & 163.45 & 0.701 & 0.552  & 272.8$_{\textcolor{BrickRed}{+14.6\%}}$ \\
 8  & 36 & \textbf{5.33} & 166.20 & 0.702 & 0.556 & 262.9$_{\textcolor{BrickRed}{+10.4\%}}$  \\
 16 & 36 & 5.49 & 168.01 & 0.700 & 0.549 & 253.0$_{\textcolor{BrickRed}{+6.3\%}}$  \\
 24 & 36 & 5.68 & 159.04 & 0.685 & 0.560 & 243.1$_{\textcolor{BrickRed}{+2.1\%}}$  \\
\midrule
 8 & 1   & 6.16 & 153.95 & 0.685 & 0.555 & 238.9$_{\textcolor{BrickRed}{+0.3\%}}$  \\
 8 & 4   & 5.47 & 165.55 & 0.700 & 0.553 & 240.8$_{\textcolor{BrickRed}{+1.1\%}}$  \\
 8 & 36  & \textbf{5.33} & 166.20 & 0.702 & 0.556 & 262.9$_{\textcolor{BrickRed}{+10.4\%}}$ \\
 8 & 100 & 5.58 & 165.61 & 0.701 & 0.546 & 307.5$_{\textcolor{BrickRed}{+29.1\%}}$ \\

\bottomrule
\end{tabular}

\label{tab:rae_ditxl_sigliP2-B_reg_depth_number_flops}
\vspace{-5pt}
\end{table}

We conduct an ablation study of diffusion registers on RAE-DiT-XL with SigLIP2-B on ImageNet 256$\times$256. The baseline model uses no diffusion registers. We study two factors: the insertion depth, defined as the transformer block from which registers are introduced, and the number of registers. In the first set of experiments, we fix the number of registers to 36 and vary the insertion depth. In the second, we fix the insertion depth to block 8 and vary the number of registers.

As shown in Tab.~\ref{tab:rae_ditxl_sigliP2-B_reg_depth_number_flops}, diffusion registers provide clear but non-monotonic gains, with a distinct sweet spot in both insertion depth and register count. When fixing the register count at 36, introducing registers too early or too late is less effective, while starting from block 8 gives the best performance. This suggests that registers are most useful when introduced in the early-to-middle part of the generator, where they can influence a substantial portion of the computation without interfering too strongly with the earliest layers. When fixing the starting block to 8, varying the number of registers shows a similar trend: very small numbers do not help, moderate numbers improve performance, the best results are obtained with 36 registers, and performance degrades at 100 registers. Overall, these results suggest that diffusion registers are most effective at an appropriate depth and capacity.

\section{Conclusion}
We study outlier tokens in Diffusion Transformers and show that they arise not only in pretrained vision encoders, but also within the diffusion model itself. Our analysis suggests that these outliers reflect degraded patch-level semantics rather than merely a few abnormally large token norms. Motivated by this insight, we introduce simple register-based interventions for both the encoder and the diffusion transformer. These interventions consistently improve training stability and generation quality in both ImageNet generation and large-scale text-to-image generation, and generalize across diverse input spaces and model variants. Overall, our results identify outlier-token control as an important ingredient for building stronger and more robust Transformer-based diffusion pipelines.

\newpage


\bibliographystyle{plainnat}
\bibliography{main}
\appendix

\section{Outlier Visualization on \textsc{SigLIP2}-So400}
\label{two-outlier}

\begin{figure}[h]
    \centering
    \includegraphics[width=0.32\linewidth]{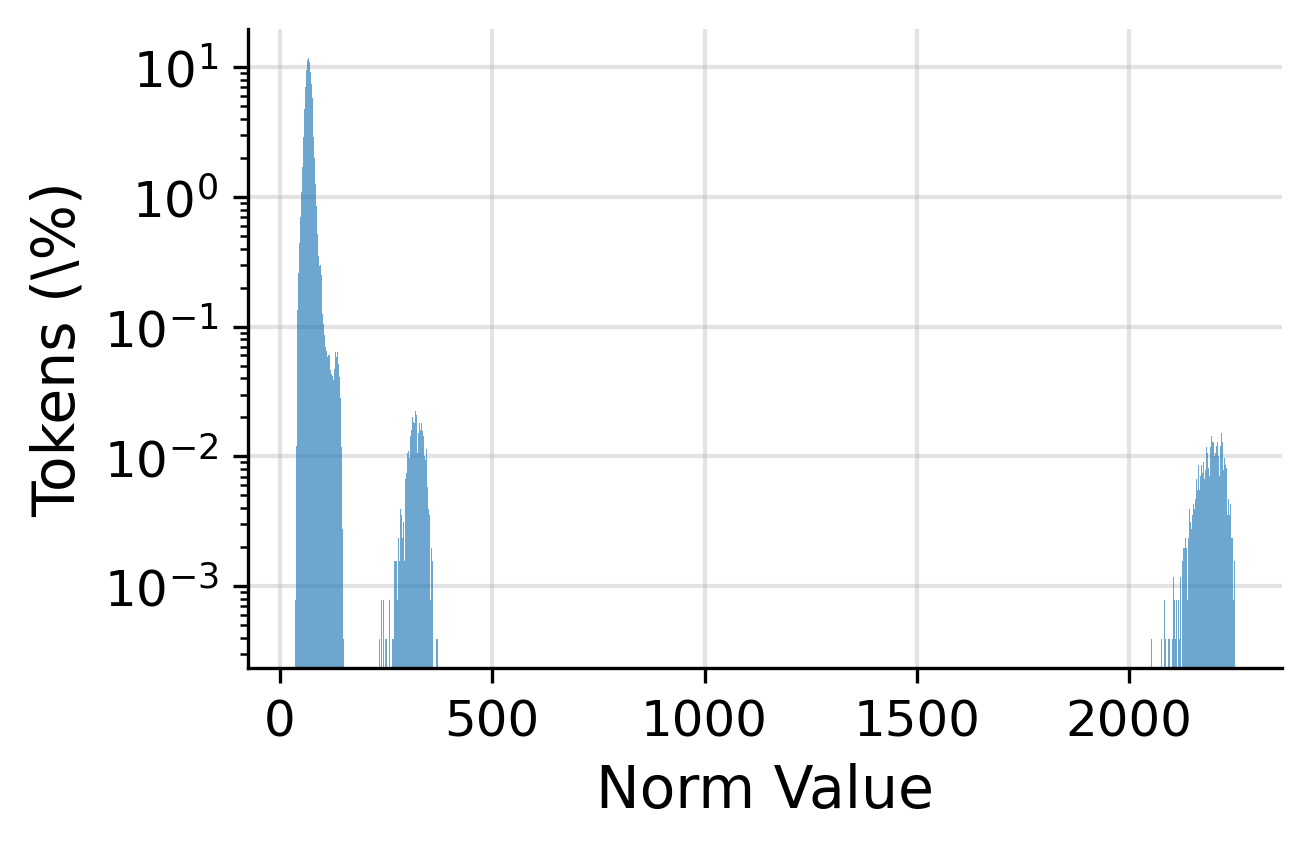}\hfill
    \includegraphics[width=0.32\linewidth]{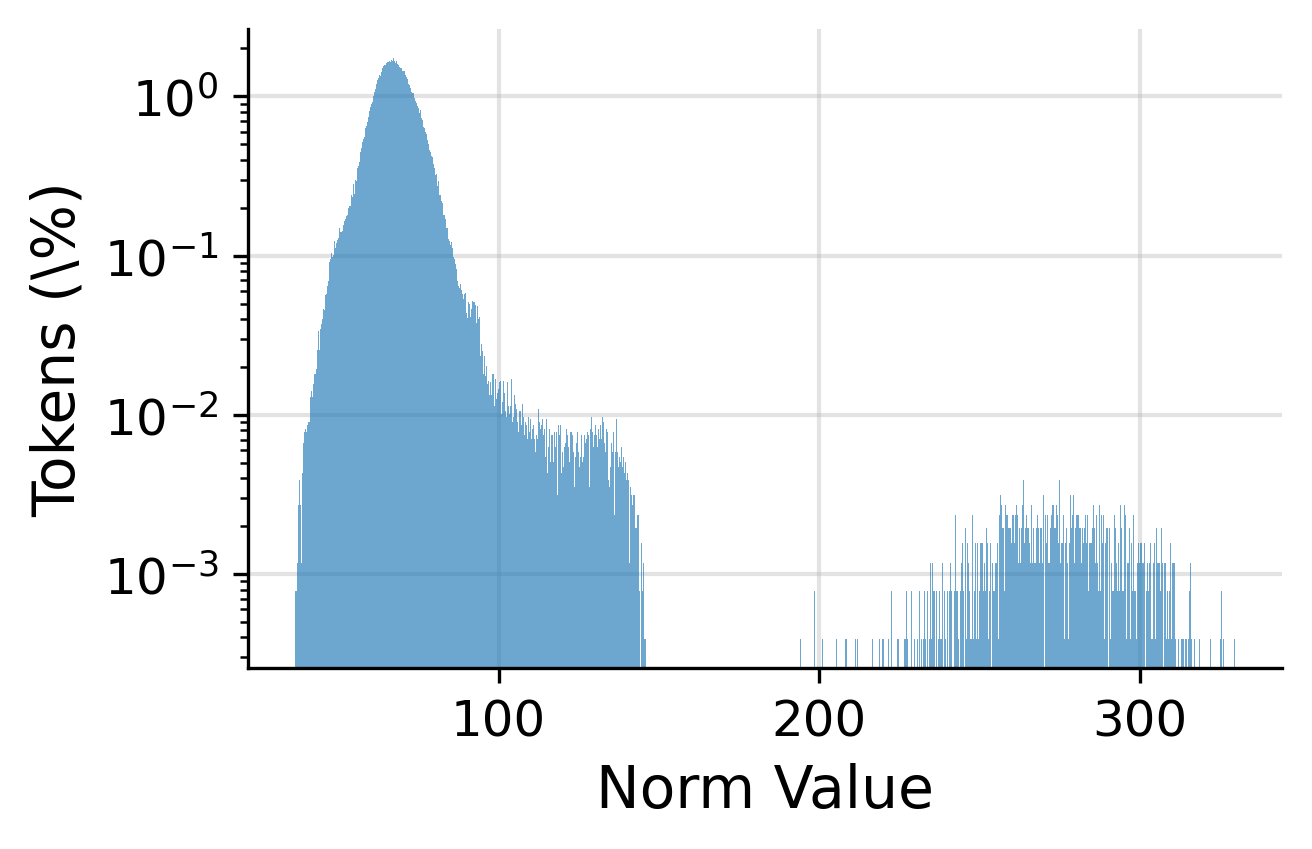}\hfill
    \includegraphics[width=0.32\linewidth]{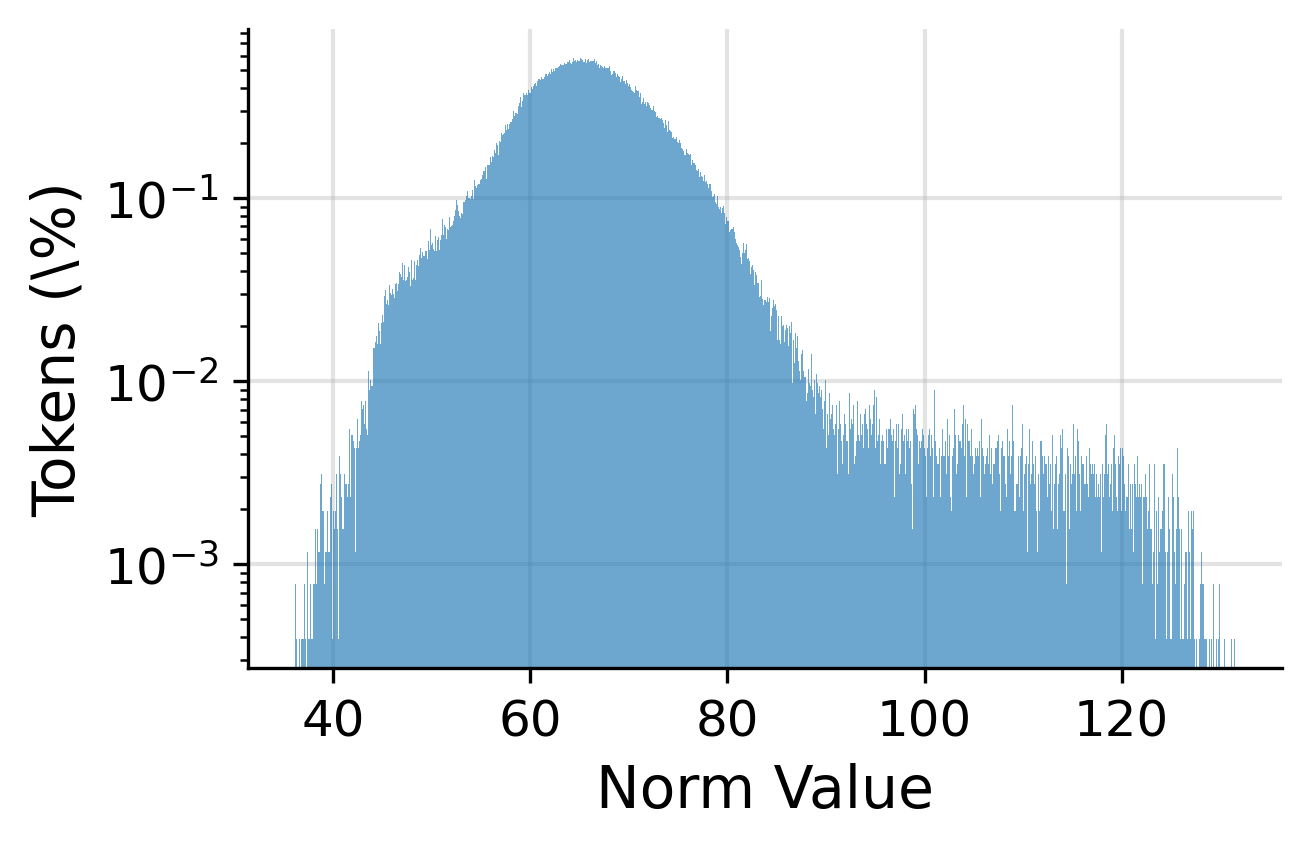}
    \caption{\textbf{Two outlier sources in the norm distribution of \textsc{SigLIP2}-So400.} We compute the $\ell_2$ norm of \textsc{SigLIP2}-So400 output features on 10k randomly selected images from the ImageNet-1K validation set.
    \textit{Left:} the original norm distribution shows two separated outlier groups.
    \textit{Middle:} applying our filtering once removes only one group, leaving the other largely intact.
    \textit{Right:} we therefore use a recursive procedure that re-localizes outlier neurons after the first pass and filters all detected outliers jointly, which effectively suppresses both groups.}
    \label{fig:siglip_so_outliers}
\end{figure}

We further analyze the norm distribution of \textsc{SigLIP2}-So400 and observe that its outliers do not form a single homogeneous group. Instead, they appear as two clearly separated clusters, indicating that a single-pass filtering strategy is insufficient. As shown in Fig.~\ref{fig:siglip_so_outliers}, filtering once removes only the more salient cluster, while the other remains visible. To address this, we adopt a recursive procedure that re-localizes outlier neurons after the first filtering pass and then jointly filters all detected outliers, allowing us to suppress both groups more effectively.

\section{More Visualization}
\begin{figure}[h]
    \centering
    \includegraphics[width=\linewidth]{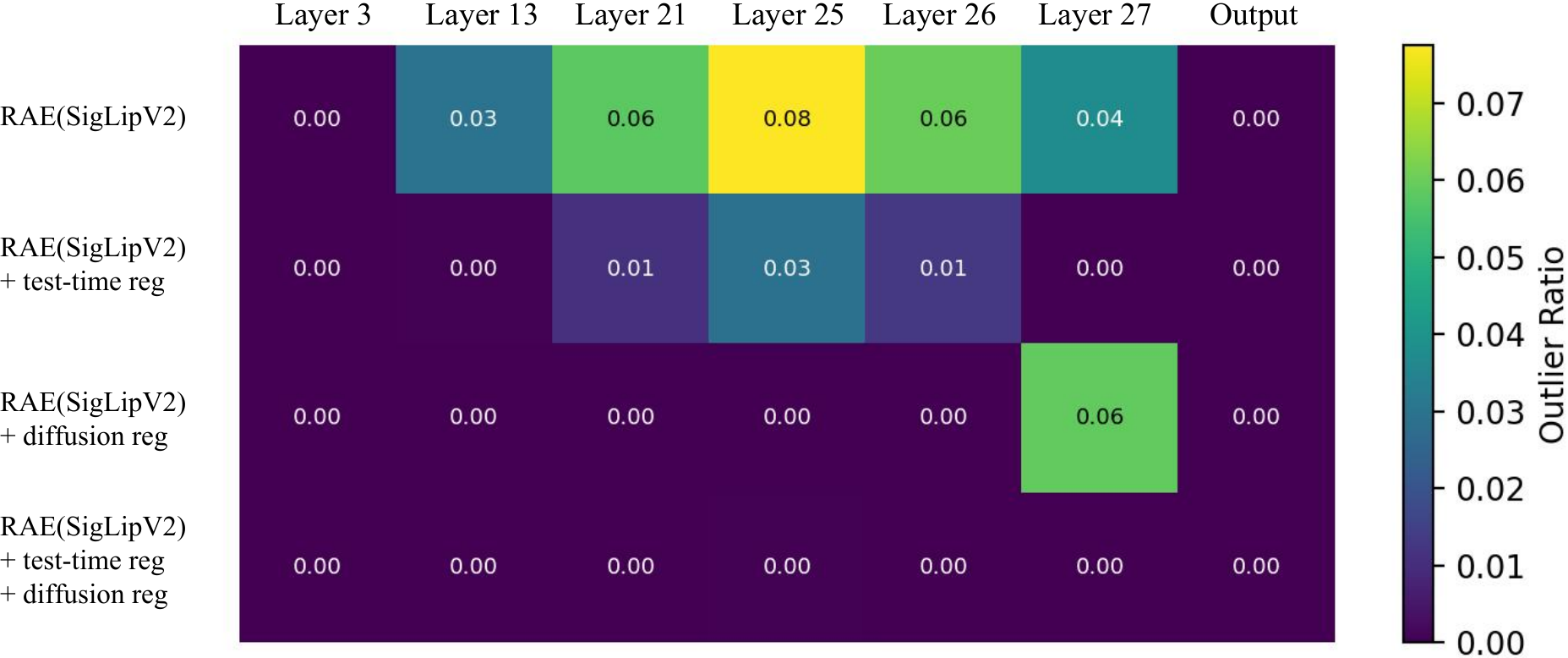}
\caption{\textbf{Quantitative measurements of outliers.} We report the fraction of outlier tokens across layers under different setups. An outlier is defined as a token whose $\ell_2$ norm exceeds $2\times$ the median token norm.}

    \label{fig:qual_map}
\end{figure}
To complement the qualitative visualization, we also provide a quantitative view of outlier behavior across layers. Figure~\ref{fig:qual_map} reports the fraction of outlier tokens under different setups, where we define an outlier as a token whose $\ell_2$ norm is greater than $2\times$ the median token norm. These results show that outlier tokens are not uniformly distributed across depth, but instead exhibit clear layer-dependent patterns that vary across settings.

\section{Retrained Decoder}
\label{retrained_decoder}
Applying test-time registers alters the encoder feature distribution. In principle, this distribution shift may require retraining the Stage-2 decoder. Interestingly, we find that the shift is mild: retraining the decoder yields no visible improvement over using the pretrained decoder.

For SigLIP2-B, we search over augmented noise scales $\sigma$, a standard knob for balancing reconstruction quality and gFID in prior work~\cite{zheng2025diffusion}. As shown in Tab.~\ref{tab:decoder-trained}, We find no setting that consistently outperforms the baseline, retraining under different $\sigma$ yields results comparable to the pretrained decoder. It suggests that test-time registers only mildly shift the encoder feature distribution seen by the decoder.

\begin{table}[t]
\centering
\small
\setlength{\tabcolsep}{16pt}
\renewcommand{\arraystretch}{1.15}
\caption{\textbf{Retraining the decoder.} Retraining the Stage-2 decoder after applying test-time registers brings negligible gains, indicating that the encoder distribution shift is mild.}
\resizebox{0.95\textwidth}{!}{%
\begin{tabular}{@{}l l c c c@{}}
\toprule
Training Strategy & Decoder& rFID$\downarrow$ & gFID$\downarrow$  &IS$\uparrow$ \\
\midrule
DiT-XL (SigLIP2-B) & Pretrained & 0.82 & 5.89 & 156.54 \\
DiT-XL (SigLIP2-B) + DSR & Pretrained & 0.82 & 4.58 & 165.99 \\
DiT-XL (SigLIP2-B) + DSR & Retrained ($\sigma{=}0.8$) & 0.58 & 6.00 & 164.20 \\
DiT-XL (SigLIP2-B) + DSR & Retrained ($\sigma{=}1.5$) & 0.67 & 5.35 & 157.59 \\
DiT-XL (SigLIP2-B) + DSR & Retrained ($\sigma{=}2.0$) & 0.75 & 5.28 & 157.98 \\
\bottomrule
\end{tabular}
}
\label{tab:decoder-trained}
\end{table}

\section{More Results on Text-to-Image Experiments}
\label{sec:scale-rae}
\begin{figure}[htbp]
    \centering
    \begin{minipage}{0.82\linewidth}
        \centering
        
        \includegraphics[width=\linewidth]{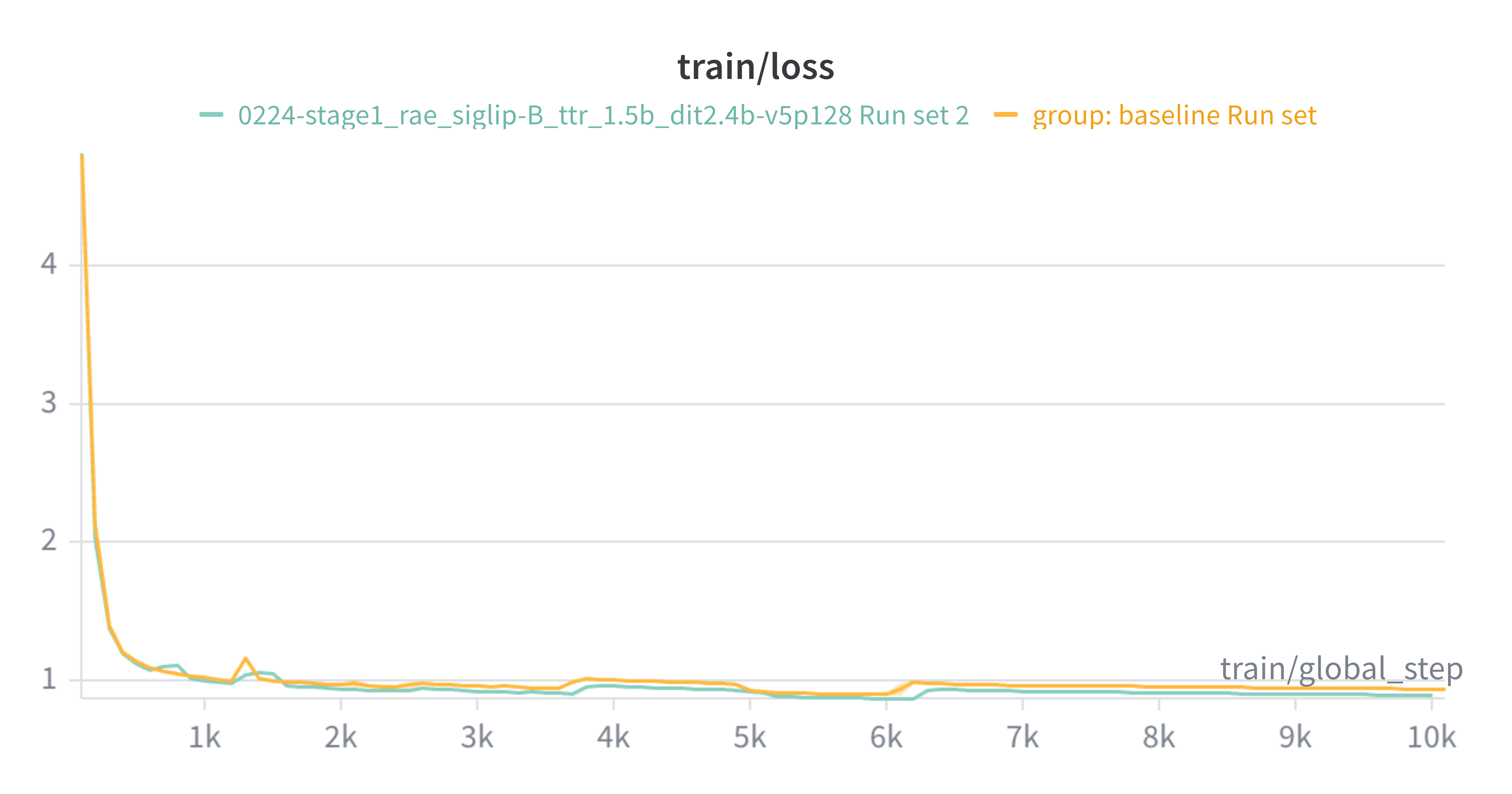}
        
        \vspace{0.1em}
        
        \begin{minipage}[t]{0.48\linewidth}
            \centering
            \includegraphics[width=\linewidth]{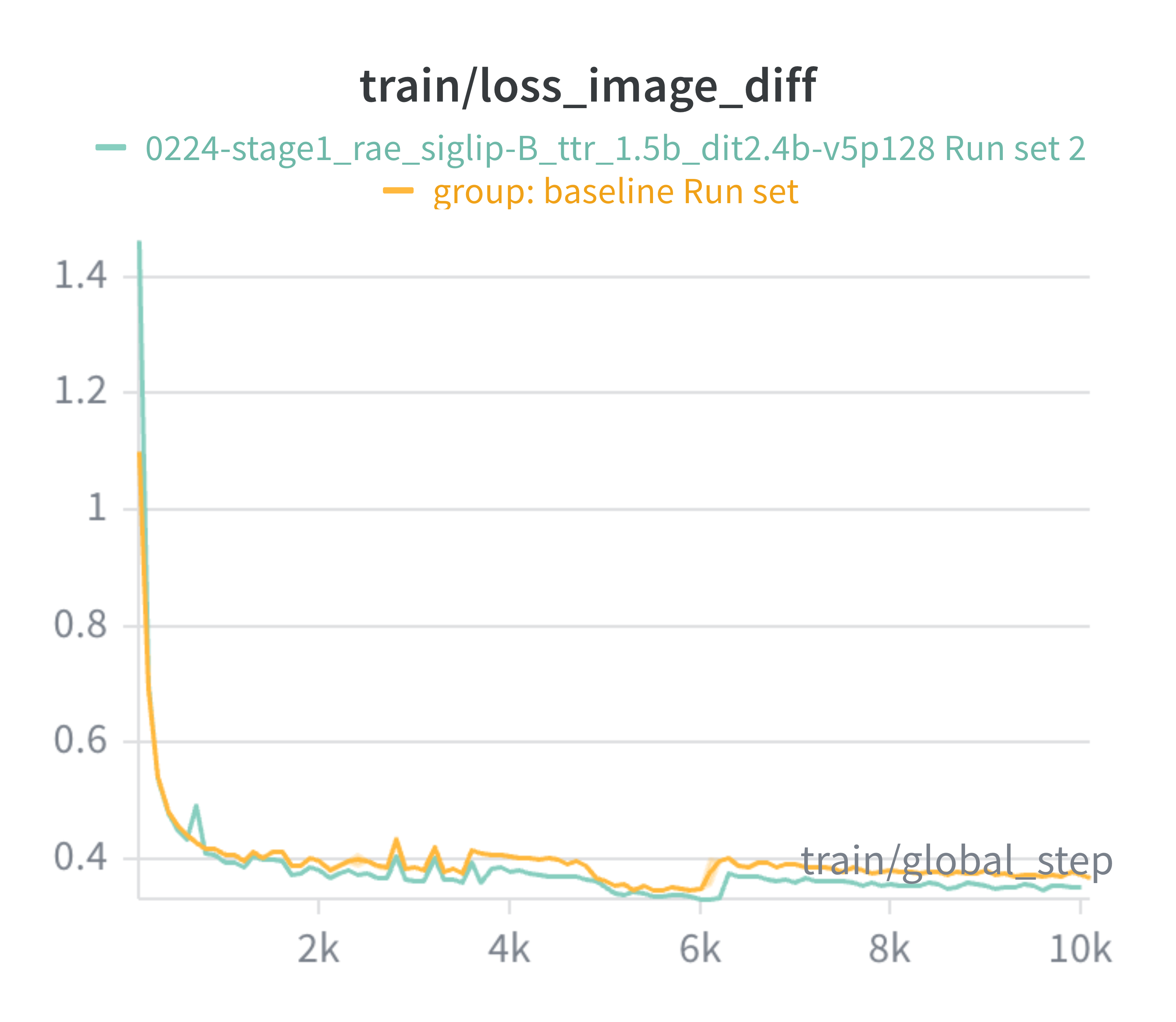}
        \end{minipage}
        \hfill
        \begin{minipage}[t]{0.48\linewidth}
            \centering
            \includegraphics[width=\linewidth]{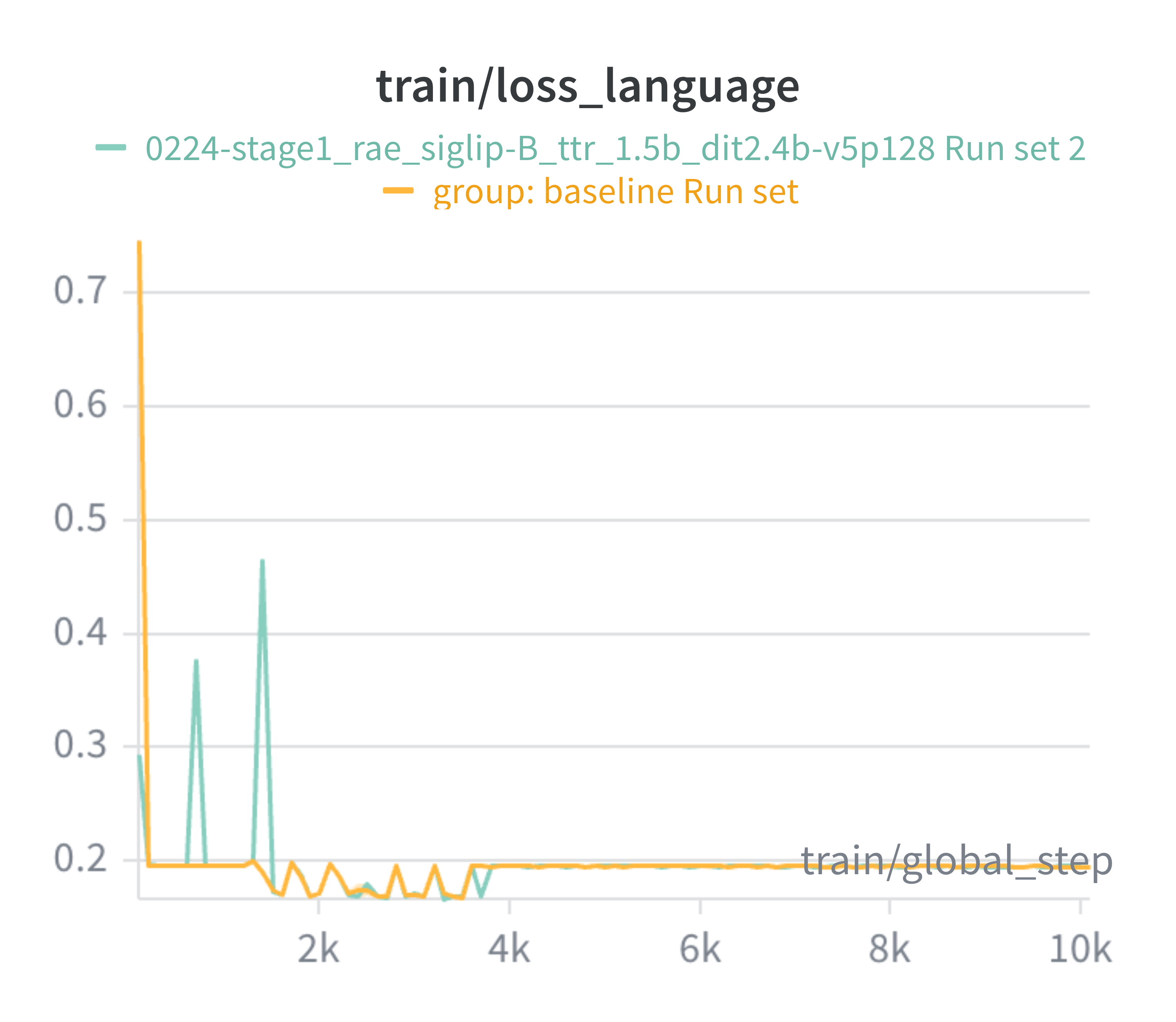}
        \end{minipage}
    \end{minipage}
    
    \caption{Training loss comparison between Scale-RAE and DSR. Top: total loss. Bottom left: image diffusion loss. Bottom right: language loss.}
    \label{fig:loss_comparison}
\end{figure}

\begin{figure}[htbp]
    \centering
    \begin{minipage}[t]{0.48\linewidth}
        \centering
        \includegraphics[width=\linewidth]{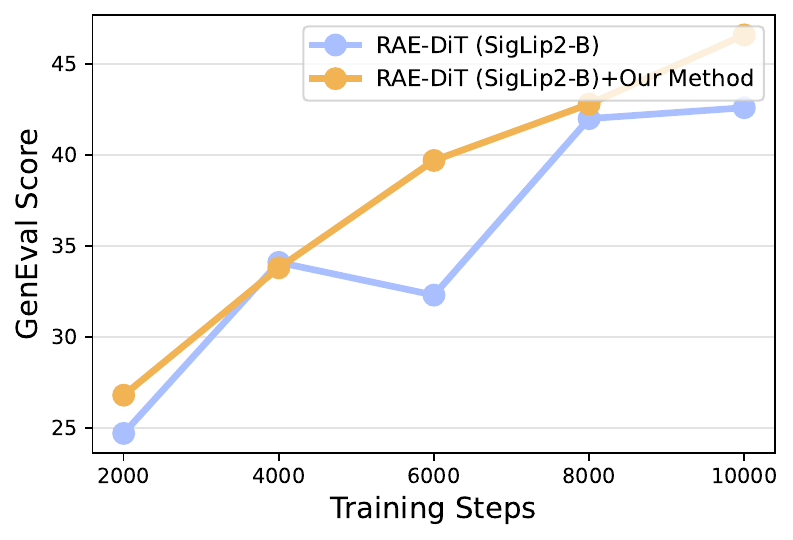}
    \end{minipage}
    \hfill
    \begin{minipage}[t]{0.48\linewidth}
        \centering
        \includegraphics[width=\linewidth]{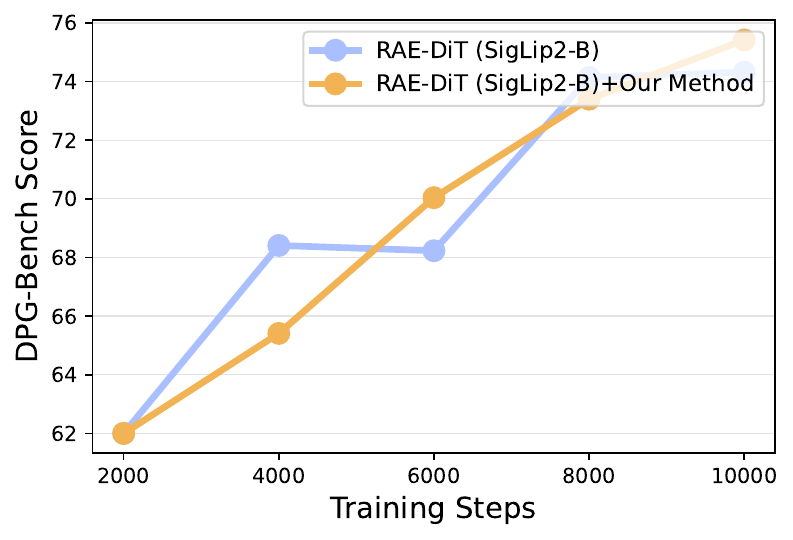}
    \end{minipage}
    \caption{Step-to-step comparison between Scale-RAE and DSR on Geneval and DPG-Bench.}
    \label{fig:geneval-and-dpgbench}
\end{figure}

\paragraph{Training configuration.}
Tab.~\ref{tab:training_config} summarizes the computational resources required and the training hyperparameter configuration for the scaling-up experiment. Following Scale-RAE~\cite{tong2026scaling}, We use SPMD
sharding together with TorchXLA to train the LLM and DiT models. Due to limited computational resources, we use only the Scale-RAE dataset, which is one quarter the size of the data used in the original Scale-RAE~\cite{tong2026scaling}, and train the model for one epoch.
\begin{table*}[t]
\centering
\small
\setlength{\tabcolsep}{6pt}
\renewcommand{\arraystretch}{1.2}
\begin{tabular}{lcccc}
\hline
 Method & \multicolumn{2}{c}{Scale-RAE} & \multicolumn{2}{c}{Scale-RAE+DSR} \\
\hline
component & LLM & DiT & LLM & DiT \\
\hline
optimizer & \multicolumn{4}{c}{AdamW} \\
learning rate schedule & \multicolumn{4}{c}{cosine w/ warmup ratio 0.0134} \\
global batch size & \multicolumn{4}{c}{2048} \\
Training epoch & \multicolumn{4}{c}{1} \\
Dataset & \multicolumn{4}{c}{Scale-RAE dataset (24.7M)} \\
max learning rate & 5e-5 & 5e-4 & 5e-5 & 5e-4 \\
optimizer betas & (0.9, 0.999) & (0.9, 0.95) & (0.9, 0.999) & (0.9, 0.95) \\
loss & autoregressive loss & diffusion loss & autoregressive loss & diffusion loss \\
model & Qwen2.5 1.5B & DiT 2.4B & Qwen2.5 1.5B & DiT 2.4B \\
Register num & - & 0 & - & 36 \\
Register adding block & - & 0 & - & 8 \\
Vision encoder & \multicolumn{4}{c}{google/siglip2-base-patch16-256} \\
Vision decoder &  \multicolumn{4}{c}{nyu-visionx/RAE-siglip2-base-p16-i256-ViTXL-n08} \\
Encoder test time register & - & No & - & Yes \\
Training device & \multicolumn{4}{c}{v5p-128} \\
TPU memory usage per chip & \multicolumn{2}{c}{22.34GiB} & \multicolumn{2}{c}{23.68GiB} \\
Training speed & \multicolumn{2}{c}{4.00s/iter} & \multicolumn{2}{c}{4.09s/iter} \\
\hline
\end{tabular}
\caption{Training configuration comparison between Scale-RAE and DSR.}
\label{tab:training_config}
\end{table*}
\paragraph{Step-to-step comparison.} We also perform a step-to-step comparison between DSR and Scale-RAE baseline. The detailed comparison is shown in Fig.~\ref{fig:geneval-and-dpgbench}. We observe consistent improvements of DSR over the Scale-RAE~\cite{tong2026scaling} baseline on GenEval~\cite{ghosh2023geneval}. On DPG-Bench~\cite{hu2024ella}, DSR shows a steadier improvement. We further compare the training losses of the baseline and DSR. As shown in Fig.~\ref{fig:loss_comparison}, the green curve corresponding to DSR is consistently lower and more stable overall, and the same trend is also visible in the diffusion loss. In contrast, the language-model loss of DSR exhibits more spikes during training, which may help explain why it sometimes underperforms the baseline on DPG-Bench, a benchmark that emphasizes following longer and more complex instructions. This suggests that additional improvements on the language side, such as mitigating the effect of outlier tokens, may further enhance performance. We leave this direction for future work.

\begin{figure}[htbp]
    \centering

    \includegraphics[width=0.18\linewidth]{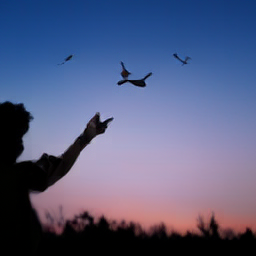}
    \hfill
    \includegraphics[width=0.18\linewidth]{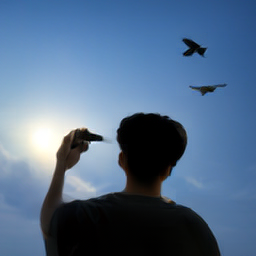}
    \hfill
    \includegraphics[width=0.18\linewidth]{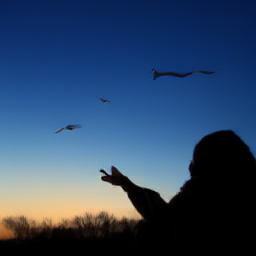}
    \hfill
    \includegraphics[width=0.18\linewidth]{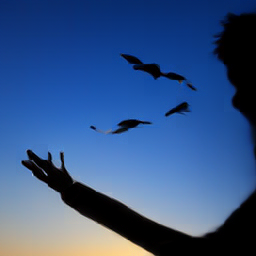}
    \hfill
    \includegraphics[width=0.18\linewidth]{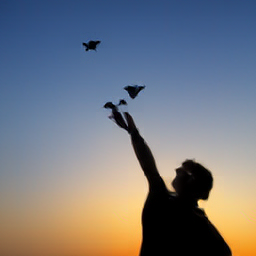}

    \vspace{0.5em}

    \includegraphics[width=0.18\linewidth]{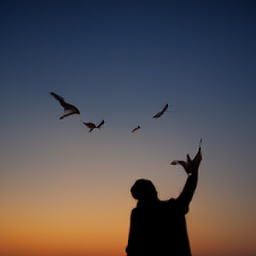}
    \hfill
    \includegraphics[width=0.18\linewidth]{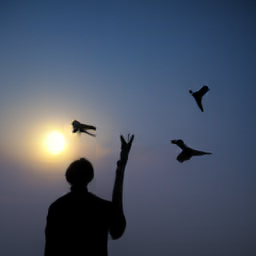}
    \hfill
    \includegraphics[width=0.18\linewidth]{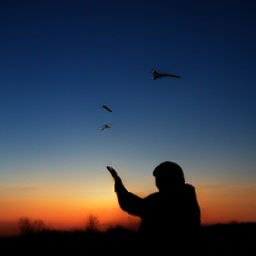}
    \hfill
    \includegraphics[width=0.18\linewidth]{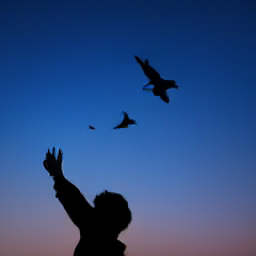}
    \hfill
    \includegraphics[width=0.18\linewidth]{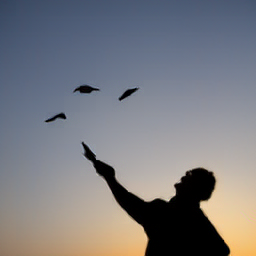}
    \caption{Scale-RAE baseline (top) vs. DSR (bottom) on DPG-Bench~\cite{hu2024ella}.  }
    \label{fig:scale1}
\end{figure}

\begin{figure}[htbp]
    \centering

    \includegraphics[width=0.18\linewidth]{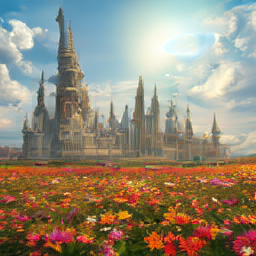}
    \hfill
    \includegraphics[width=0.18\linewidth]{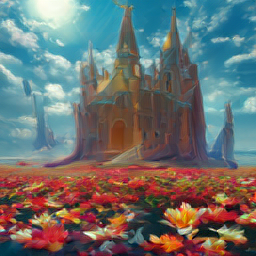}
    \hfill
    \includegraphics[width=0.18\linewidth]{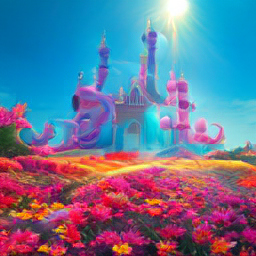}
    \hfill
    \includegraphics[width=0.18\linewidth]{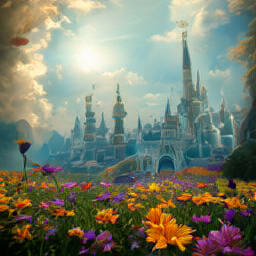}
    \hfill
    \includegraphics[width=0.18\linewidth]{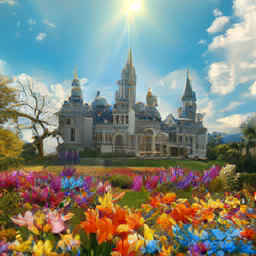}

    \vspace{0.5em}

    \includegraphics[width=0.18\linewidth]{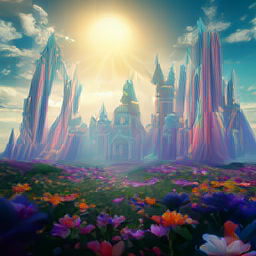}
    \hfill
    \includegraphics[width=0.18\linewidth]{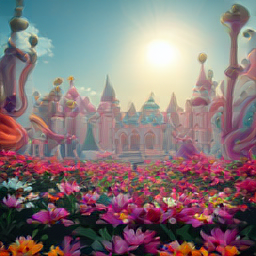}
    \hfill
    \includegraphics[width=0.18\linewidth]{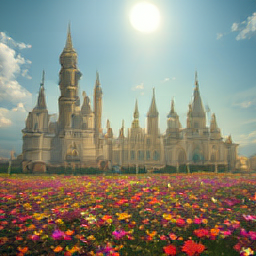}
    \hfill
    \includegraphics[width=0.18\linewidth]{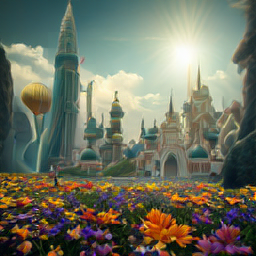}
    \hfill
    \includegraphics[width=0.18\linewidth]{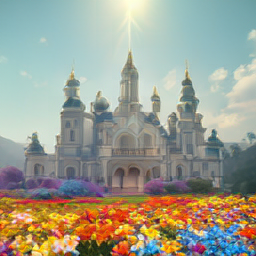}
    \caption{Scale-RAE baseline (top) vs. DSR (bottom) on DPG-Bench~\cite{hu2024ella}. }
    \label{fig:scale2}
\end{figure}

\begin{figure}[htbp]
    \centering

    \includegraphics[width=0.18\linewidth]{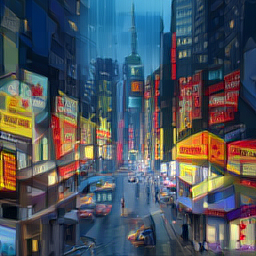}
    \hfill
    \includegraphics[width=0.18\linewidth]{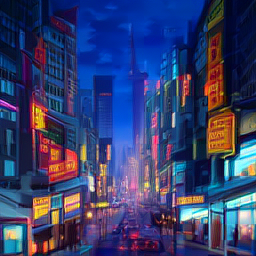}
    \hfill
    \includegraphics[width=0.18\linewidth]{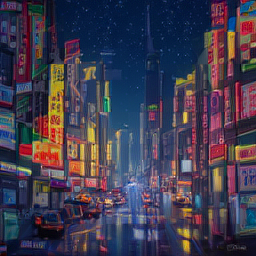}
    \hfill
    \includegraphics[width=0.18\linewidth]{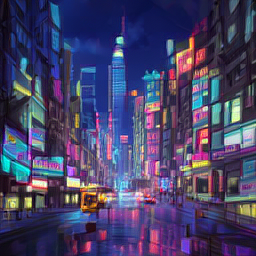}
    \hfill
    \includegraphics[width=0.18\linewidth]{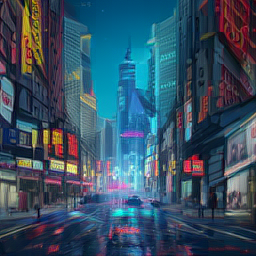}

    \vspace{0.5em}

    \includegraphics[width=0.18\linewidth]{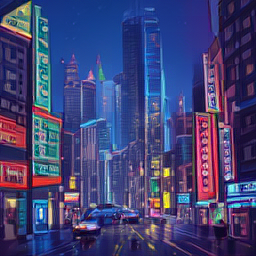}
    \hfill
    \includegraphics[width=0.18\linewidth]{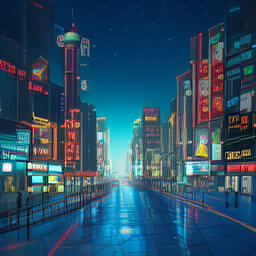}
    \hfill
    \includegraphics[width=0.18\linewidth]{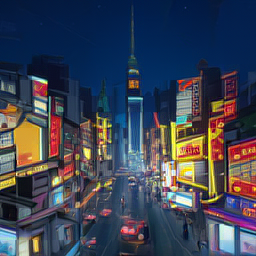}
    \hfill
    \includegraphics[width=0.18\linewidth]{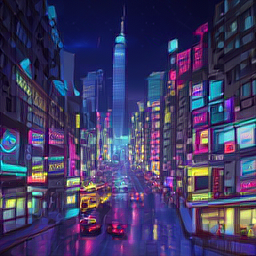}
    \hfill
    \includegraphics[width=0.18\linewidth]{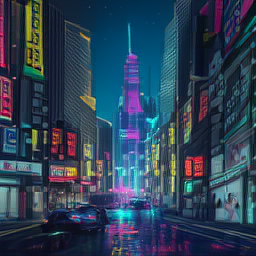}
    \caption{Scale-RAE baseline (top) vs. DSR (bottom) on DPG-Bench~\cite{hu2024ella}. }
    \label{fig:scale3}
\end{figure}

\begin{figure}[htbp]
    \centering

    \includegraphics[width=0.18\linewidth]{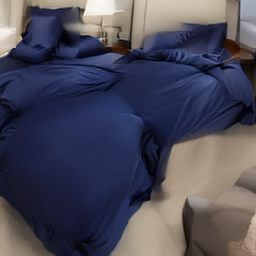}
    \hfill
    \includegraphics[width=0.18\linewidth]{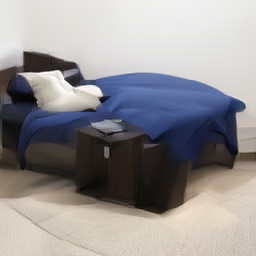}
    \hfill
    \includegraphics[width=0.18\linewidth]{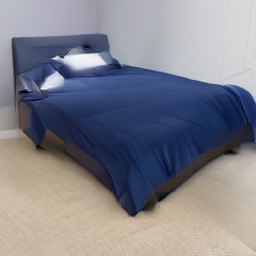}
    \hfill
    \includegraphics[width=0.18\linewidth]{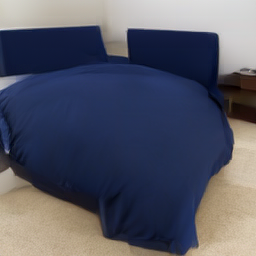}
    \hfill
    \includegraphics[width=0.18\linewidth]{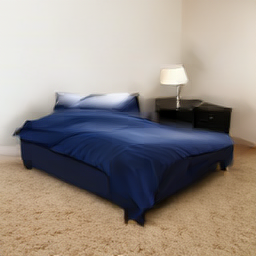}

    \vspace{0.5em}

    \includegraphics[width=0.18\linewidth]{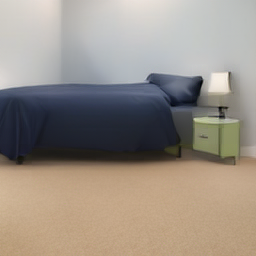}
    \hfill
    \includegraphics[width=0.18\linewidth]{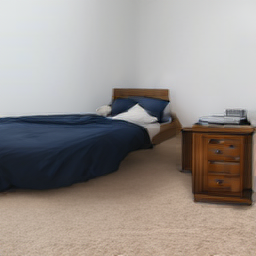}
    \hfill
    \includegraphics[width=0.18\linewidth]{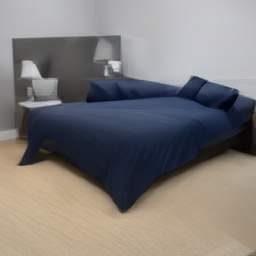}
    \hfill
    \includegraphics[width=0.18\linewidth]{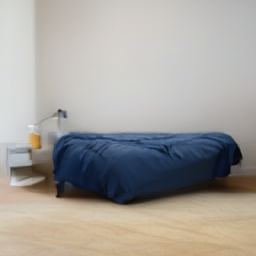}
    \hfill
    \includegraphics[width=0.18\linewidth]{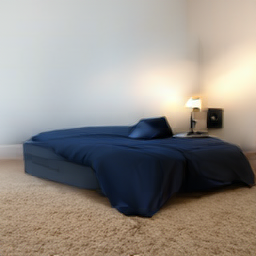}
    \caption{Scale-RAE baseline (top) vs. DSR (bottom) on DPG-Bench~\cite{hu2024ella}. }
    \label{fig:scale4}
\end{figure}

\begin{figure}[htbp]
    \centering

    \includegraphics[width=0.18\linewidth]{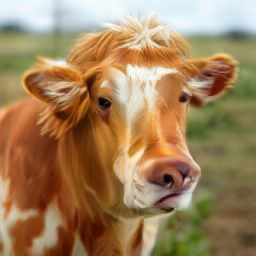}
    \hfill
    \includegraphics[width=0.18\linewidth]{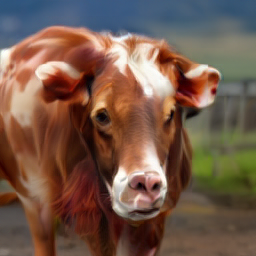}
    \hfill
    \includegraphics[width=0.18\linewidth]{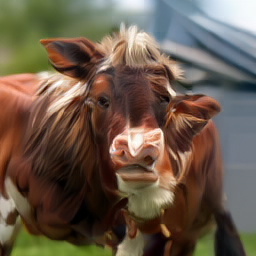}
    \hfill
    \includegraphics[width=0.18\linewidth]{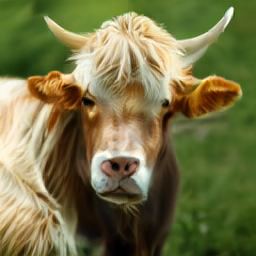}
    \hfill
    \includegraphics[width=0.18\linewidth]{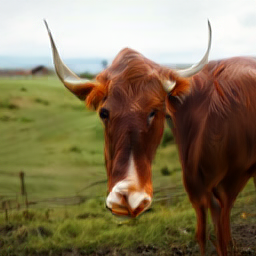}

    \vspace{0.5em}

    \includegraphics[width=0.18\linewidth]{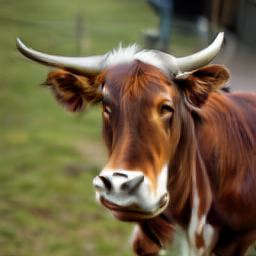}
    \hfill
    \includegraphics[width=0.18\linewidth]{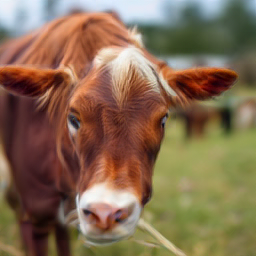}
    \hfill
    \includegraphics[width=0.18\linewidth]{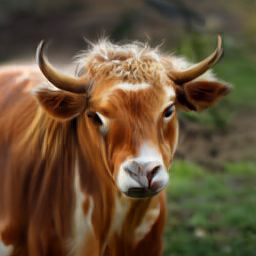}
    \hfill
    \includegraphics[width=0.18\linewidth]{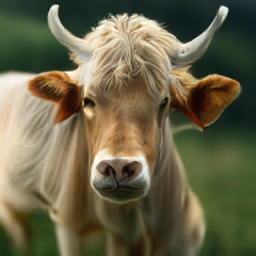}
    \hfill
    \includegraphics[width=0.18\linewidth]{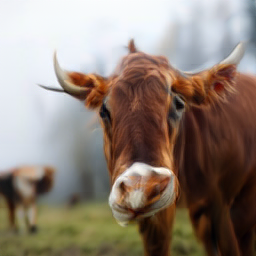}
    \caption{Scale-RAE baseline (top) vs. DSR (bottom) on GenEval~\cite{ghosh2023geneval}. }
    \label{fig:scale5}
\end{figure}

\begin{figure}[htbp]
    \centering

    \includegraphics[width=0.18\linewidth]{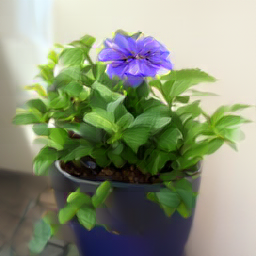}
    \hfill
    \includegraphics[width=0.18\linewidth]{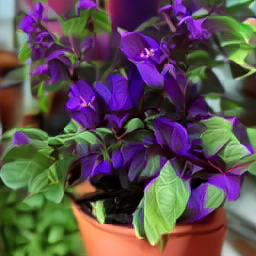}
    \hfill
    \includegraphics[width=0.18\linewidth]{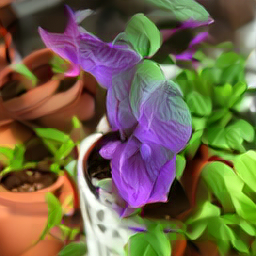}
    \hfill
    \includegraphics[width=0.18\linewidth]{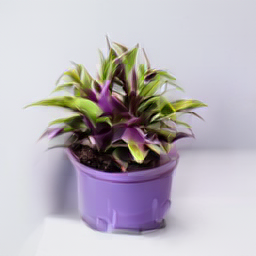}
    \hfill
    \includegraphics[width=0.18\linewidth]{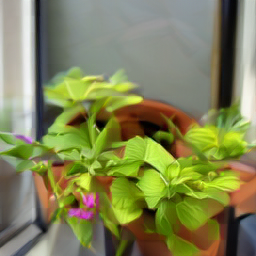}

    \vspace{0.5em}

    \includegraphics[width=0.18\linewidth]{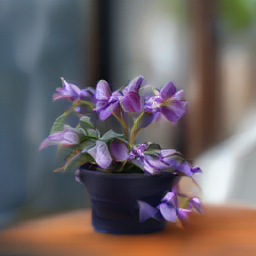}
    \hfill
    \includegraphics[width=0.18\linewidth]{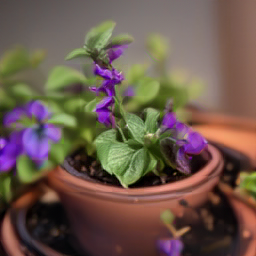}
    \hfill
    \includegraphics[width=0.18\linewidth]{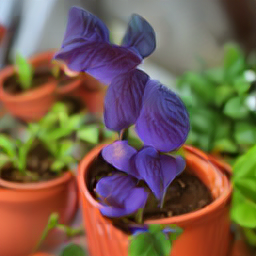}
    \hfill
    \includegraphics[width=0.18\linewidth]{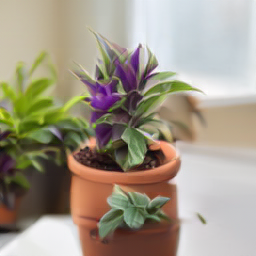}
    \hfill
    \includegraphics[width=0.18\linewidth]{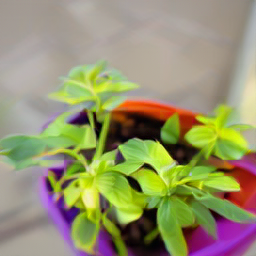}
    \caption{Scale-RAE baseline (top) vs. DSR (bottom) on GenEval~\cite{ghosh2023geneval}. }
    \label{fig:scale6}
\end{figure}

\begin{figure}[htbp]
    \centering

    \includegraphics[width=0.18\linewidth]{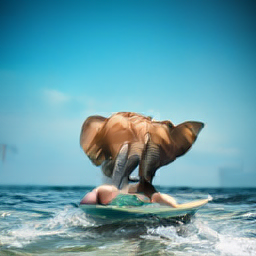}
    \hfill
    \includegraphics[width=0.18\linewidth]{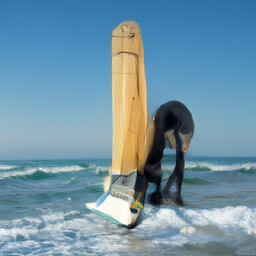}
    \hfill
    \includegraphics[width=0.18\linewidth]{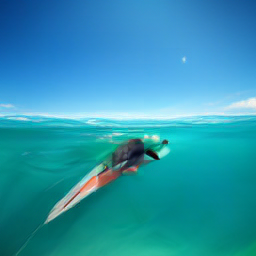}
    \hfill
    \includegraphics[width=0.18\linewidth]{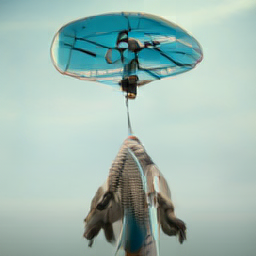}
    \hfill
    \includegraphics[width=0.18\linewidth]{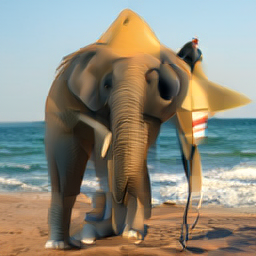}

    \vspace{0.5em}

    \includegraphics[width=0.18\linewidth]{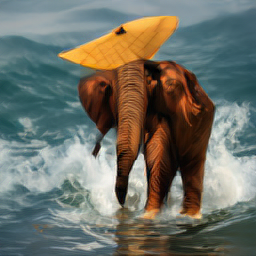}
    \hfill
    \includegraphics[width=0.18\linewidth]{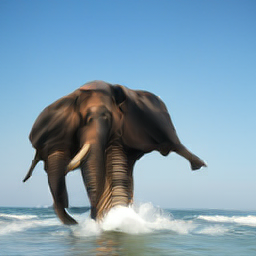}
    \hfill
    \includegraphics[width=0.18\linewidth]{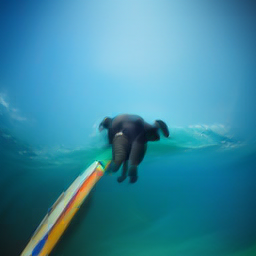}
    \hfill
    \includegraphics[width=0.18\linewidth]{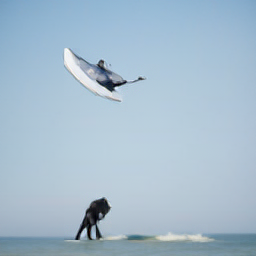}
    \hfill
    \includegraphics[width=0.18\linewidth]{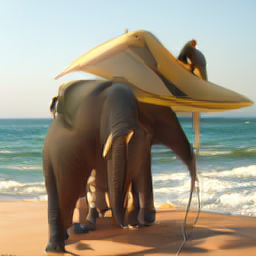}
    \caption{Scale-RAE baseline (top) vs. DSR (bottom) on GenEval~\cite{ghosh2023geneval}. }
    \label{fig:scale7}
\end{figure}

\begin{figure}[htbp]
    \centering

    \includegraphics[width=0.18\linewidth]{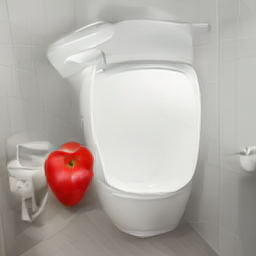}
    \hfill
    \includegraphics[width=0.18\linewidth]{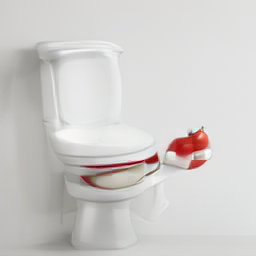}
    \hfill
    \includegraphics[width=0.18\linewidth]{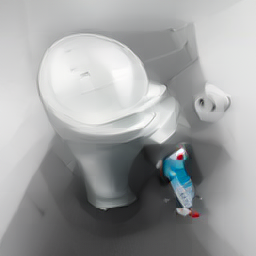}
    \hfill
    \includegraphics[width=0.18\linewidth]{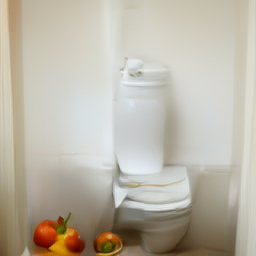}
    \hfill
    \includegraphics[width=0.18\linewidth]{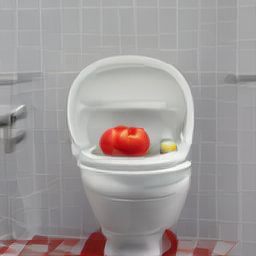}

    \vspace{0.5em}

    \includegraphics[width=0.18\linewidth]{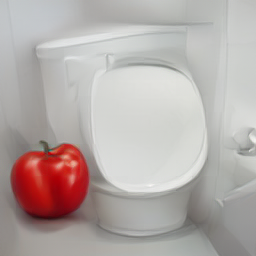}
    \hfill
    \includegraphics[width=0.18\linewidth]{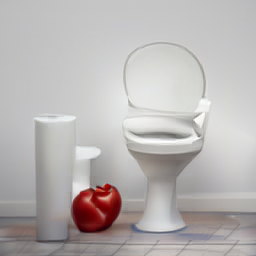}
    \hfill
    \includegraphics[width=0.18\linewidth]{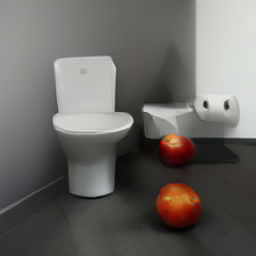}
    \hfill
    \includegraphics[width=0.18\linewidth]{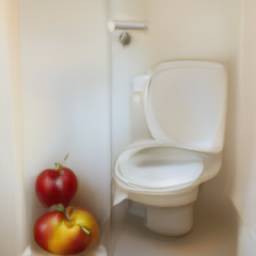}
    \hfill
    \includegraphics[width=0.18\linewidth]{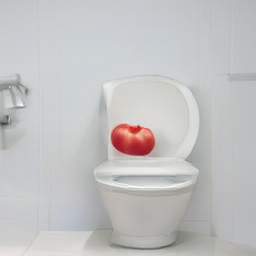}
    \caption{Scale-RAE baseline (top) vs. DSR (bottom) on GenEval~\cite{ghosh2023geneval}. }
    \label{fig:scale8}
\end{figure}

\paragraph{Extended samples.}
We provide extended samples from Scale-RAE~\cite{tong2026scaling} and DSR on DPG-Bench~\cite{hu2024ella} and Geneval~\cite{ghosh2023geneval} (Fig.~\ref{fig:scale1}, Fig.~\ref{fig:scale2}, Fig.~\ref{fig:scale3}, Fig.~\ref{fig:scale4}, Fig.~\ref{fig:scale5}, Fig~\ref{fig:scale6},
Fig~\ref{fig:scale7},Fig~\ref{fig:scale8}). The prompts for generated images are:
\begin{itemize}
    \item During the twilight hour, an individual can be seen extending an arm towards the sky, pointing at a trio of wild birds gliding through the rich deep blue of the early evening sky. The birds' silhouettes contrast distinctly against the fading light, their wings spread wide as they soar. The person is silhouetted against the dusky sky, creating a peaceful scene of human connection with nature.
    \item An expansive palace constructed from iridescent materials that shimmer with hues reminiscent of a vivid, Slime-like substance, majestically stands at the heart of a fantastical realm. Its towers twist skyward, defying conventional architecture with their organic, flowing forms. In the foreground, a field of exotic flowers blooms, each petal displaying an array of otherworldly colors that could have been plucked from a Lovecraftian spectrum, while overhead, a radiant sun bathes the surreal landscape in brilliant light.
    \item a highly intricate and vibrant cityscape that reflects a fusion of Moebius's imaginative design and Makoto Shinkai's detailed animation style. The streets are aglow with neon signs in a kaleidoscope of colors, casting reflections on the glossy, rain-slicked pavements. Towering skyscrapers with glowing windows rise towards a starless night sky, as the artwork garners significant attention and praise on ArtStation.
    \item Adjacent to each other in a room, a large rectangular bed draped in a navy-blue comforter sits parallel to a square-shaped nightstand with a matte finish. The nightstand holds an angular lamp and a small stack of hardcover books. The two pieces of furniture are positioned on a plush beige carpet that covers the majority of the floor space.
    \item a photo of a cow.
    \item a photo of a purple potted plant.
    \item a photo of an elephant below a surfboard.
    \item a photo of a white toilet and a red apple.
    
\end{itemize}

\end{document}